\documentclass[nohyperref]{article}

\usepackage{todonotes}
\usepackage{amsmath}
\usepackage{amsthm}
\usepackage{amssymb}
\usepackage{amsfonts} 
\usepackage{mathrsfs}
\usepackage{graphicx}
\usepackage{epstopdf}
\usepackage{nicefrac}
\usepackage{enumitem}
\usepackage{multicol}
\usepackage{bm}
\usepackage{bbm}
\usepackage{color,soul}
\usepackage{mathtools}
\usepackage{microtype}
\usepackage{graphicx}
\usepackage{subfigure}
\usepackage{booktabs} 
\usepackage{hyperref}

\usepackage{hyperref}

\usepackage[accepted]{icml2022}

\icmltitlerunning{Faster Algorithms for Learning Convex Functions}

\newtheorem{lemma}{Lemma}

\newtheorem{proposition}{Proposition}
\newtheorem{theorem}{Theorem}
\newtheorem{corollary}{Corollary}
\newtheoremstyle{named}{}{}{\itshape}{}{\bfseries}{.}{.5em}{#1 #3}
\theoremstyle{named}
\newtheorem*{namthm*}{Theorem}
\usepackage{mathtools}

\DeclareMathOperator{\sign}{sign}
\newcommand{\ma}{{\bf A}}
\newcommand{\mb}{{\bf B}}

\usepackage{algorithmic, algorithm}

\begin{document}

\twocolumn[
\icmltitle{Faster Algorithms for Learning Convex Functions}



\icmlsetsymbol{equal}{*}

\begin{icmlauthorlist}
\icmlauthor{Ali Siahkamari}{equal,BU}
\icmlauthor{Durmus Alp Emre Acar}{equal,BU}
\icmlauthor{Christopher Liao}{BU}
\icmlauthor{Kelly Geyer}{BU}
\icmlauthor{Venkatesh Saligrama}{BU}
\icmlauthor{Brian Kulis}{BU}
\end{icmlauthorlist}


\icmlaffiliation{BU}{Boston University, Boston, MA}

\icmlcorrespondingauthor{Ali Siahkamari}{siaa@bu.edu}

\icmlkeywords{Machine Learning, ICML}

\vskip 0.3in
]



\printAffiliationsAndNotice{\icmlEqualContribution} 

\begin{abstract}
The task of approximating an arbitrary convex function arises in several learning problems such as convex regression, learning with a difference of convex (DC) functions, and learning Bregman or $f$-divergences. In this paper, we develop and analyze an approach for solving a broad range of convex function learning problems that is faster than state-of-the-art approaches.  Our approach is based on a 2-block ADMM method where each block can be computed in closed form.  For the task of convex Lipschitz regression, we establish that our proposed algorithm converges with iteration complexity of $ O(n\sqrt{d}/\epsilon)$   for a dataset $\bm X \in \mathbb R^{n\times d}$ and $\epsilon > 0$. Combined with per-iteration computation complexity, our method converges with the rate $O(n^3 d^{1.5}/\epsilon+n^2 d^{2.5}/\epsilon+n d^3/\epsilon)$. This new rate improves the state of the art  rate of $O(n^5d^2/\epsilon)$ if $d = o( n^4)$.  Further we provide similar solvers for DC regression and Bregman divergence learning.  Unlike previous approaches, our method is amenable to the use of GPUs.   We demonstrate on regression and metric learning experiments that our approach is over 100 times faster than existing approaches on some data sets, and produces results that are comparable to state of the art.
\end{abstract}

\section{Introduction}
The importance of convex functions in machine learning is undisputed.  However, while most applications of convexity in machine learning involve fixed convex functions, an emerging trend in the field has focused on the problem of \textit{learning} convex functions for a particular task.  In this setting, data or supervision is used to tailor a convex function for some end goal.  Recent machine learning examples include learning with a difference of convex (DC) functions~\citep{siahkamari2020piecewise}, learning input convex neural networks for tasks in reinforcement learning or vision~\citep{amos2017input}, and learning divergences between points or distributions via learning the underlying convex function in a Bregman divergence~\citep{siahkamari2019learning} or $f$-divergence~\citep{zhang2020f} for problems such as data generation, metric learning, and others.

In fact, work in convex function learning at least dates back to methods developed for the problem of \textit{convex regression}, an important learning problem that is used commonly in econometrics and engineering for modeling demand, utility and production curves \citep{afriat1967construction, varian1982nonparametric}.  
Convex regression is the problem of estimating a convex function when receiving a dataset $\{(\bm x_i, y_i)\}_{i=1}^n$, where $\bm x_i \in \mathbb{R}^{d}$ are predictors and $y_i$ are continuous responses. Consider 
\begin{align} \label{equ:convexclass}
    \mathcal{F} \triangleq \{f: \mathbb{R}^d \to \mathbb{R}\ | \ f \text{ is convex}\},
\end{align}
as the class of all convex functions over $\mathbb{R}^d$. 
Then an estimator is proposed by minimizing the squared error between observations $\bm{x}_i$ and responses $y_i$,
\begin{align}\label{equ:srm}
\hat f \triangleq \arg \min_{f\in \mathcal{F}} \frac{1}{n}\sum_{i=1}^n(y_i -f(\bm{x}_i))^2 + \lambda \|f\|,
\end{align}
where $\|f\|$ is some penalty term.
A key insight about this problem is that although Eq. \eqref{equ:srm} is an infinite dimensional minimization problem, \citet{boyd2004convex} shows it can be solved as a convex optimization problem since the optimal solution can be shown to be piece-wise linear. 
Furthermore, generalization bounds for this problem have recently been developed in \citep{balazs2016convex, siahkamari2020piecewise} using certain classes for $\|f\|$.

Many other convex function learning problems have a similar structure to them, and also yield resulting optimization problems that can be tractably solved.  However, a key downside is that the computational complexity is often prohibitive for real-world applications, particularly due to the non-parametric nature of the resulting optimization problems.  For instance, the state-of-the-art solver for convex regression is based on interior point methods and has a complexity of $O(n^5d^2/\epsilon)$ for a given accuracy $\epsilon$.  

Thus, our goal in this paper is to study a class of methods that yields provably faster convergence for a number of different convex function learning problems.  
Our main insight is that many convex function learning problems may be expressed as optimization problems that can be solved with a 2-block ADMM algorithm such that each of the two blocks can be computed \textit{in closed form}.  For convex regression, the resulting method (which we call FCR, for Faster Convex Regression) is guaranteed to converge with computational complexity of $O(n^3 d^{1.5}/\epsilon+n^2 d^{2.5}/\epsilon+n d^3/\epsilon)$ for a dataset $\bm X \in \mathbb R^{n\times d}$ and $\epsilon>0$. 
This new rate improves the state of the art $O(n^5d^2/\epsilon)$ in \cite{balazs2016convex} available via interior point methods when $d=o(n^4)$. 
In addition to an improved convergence rate, FCR makes several contributions.
Firstly, FCR is stand-alone and does not need any optimization software.
FCR is based on tensor computations and can easily be implemented on GPUs.
Furthermore, FCR can include regularization terms in the context of non-parametric convex regression. 

We extend FCR solver for problems beyond convex regression.
In particular, we provide 2-block ADMM solvers and present empirical comparisons for DC regression  \citep{siahkamari2020piecewise} and Bregman divergence learning \citep{siahkamari2019learning}. Finally, we demonstrate empirical results showing that our approach yields significantly more scalable convex learning methods, resulting in speedups of over 100 times on some problems.

    
    
    
\textbf{Notation.}
We generally denote scalars as lower case letters, vectors as lower case bold letters, and matrices as upper case bold letters.
For a dataset $\bm X \in \mathbb{R}^{n \times d}$, $n$ represents the number of observations and $d$ is the number of covariates.
We define $x^+ = \max\{x,0\}, x^- =  \max\{-x,0\}$ and $\times$ as an outer product. We define $[n]$ to be the set of integers $\{1,\dots, n\}$.  By $\|\nabla^* f(\bm x)\|$ and $|\partial^*_{x_l}f(\bm x)| $ we denote the largest subgradient and the largest partial sub-derivative. \\

\subsection{Connections to Existing Methodologies}
Convex regression has been extensively studied over the last two decades. \citet{boyd2004convex} introduce the non-parametric convex regression problem Eq. \eqref{equ:srm}.
\citet{balazs2016convex} show that convex Lipschitz regression requires regularization in order to have generalization guarantees.
In particular, it was shown that the penalty term $\|f\|=\sup_{\bm x}\|\nabla^* f(\bm x)\|$ yields a generalization error of $O(n^{-2/d}\log n)$.

%
\citet{balazs2016convex, mazumder2019computational} provide ADMM  solvers for  the convex regression problem; however, solutions have more than two blocks and are not guaranteed to converge in all circumstances. 
More recently, \citet{chen2020multivariate} provide an active-set type solver and \citet{bertsimas2021sparse} provide a delayed constraint generation algorithm which have favorable scalablity. 
They use a penalty of $\|f\| = \sum_i \|\nabla^* f(\bm x_i)\|_2^2$, which makes the loss function in Eq. \eqref{equ:srm} strongly convex and easier to solve. 
However, no theoretical generalization bound is known when using this new penalty term. 
\citet{siahkamari2020piecewise} use $\|f\|=\sup_{\bm x}(\|\nabla^* f(\bm x)\|_1)$ for learning a difference of convex functions where they provide a multi-block ADMM solver for this problem. 

On the other hand, \citet{ghosh2019max} and \citet{kim2021max} study the parametric class of max-linear functions $\mathcal{F}_k=\{f : \mathbb{R}^d \to \mathbb{R} \ | \ f(\bm x)= \max_{j=1}^k \langle \bm a_j, \bm x \rangle + b_j \}$ for $k < n$. They provide convex programming and alternating algorithms. 
The downside of these methods is the assumption that $\bm x_i$ are i.i.d samples from a normal distribution and furthermore that each feature $x_{i,l}$ is independent. 
These assumptions are needed for their algorithm to converge in theory. 
We note that this parametric class $\mathcal{F}_k$ if $k\geq n$, is the same as $\mathcal{F}$, when used in convex regression minimization problem Eq.~(\ref{equ:srm}).

Our methodology is most similar to \citet{balazs2016convex} and is in the context of learning theory where we do not require any distributional assumption on $\bm x_i$ and only require them be i.i.d. and bounded. We further need to know a bound on $f$ but not on $\|f\|$.

\section{Convex Regression with Lasso Penalty}

Suppose we are given a dataset $\{(y_i,\bm x_i)\}_{i=1}^n$ where $\bm x_i \in \mathbb{R}^{d}$ are predictors, assumed drawn i.i.d. from a distribution $P_X$ supported on a compact domain $\Omega \subset \{\|\bm x\|_\infty \le R\},$ with $n\ge d$, and $y_i$ are responses such that $y_i = f(\bm x_i) + \varepsilon_i$ for centered, independent random noise $\varepsilon_i$. Assume $|\varepsilon_i|$ and $|f(\cdot)|$ both are bounded by $M$. Furthermore, $f$ is convex and Lipschitz. We propose to solve the penalized convex regression problem in Eq.~(\ref{equ:srm}) with the penalty term $\|f\|\triangleq \sum_{l=1}^d \sup_{\bm x} | \partial_{x_l}^* f(\bm x)|$.  This results in the following convex optimization problem:
\begin{align}\label{program:convex}
&\qquad \min_{\hat{y}_i , \bm a_i}\frac{1}{n} \sum_{i=1}^n  (\hat y_i - y_i)^2 + \lambda \sum_{l=1}^d \max_{i=1}^n | a_{i,l}|  \\
&\textrm{s.t. } 
\hat y_i -  \hat y_j - \langle \bm a_i, \bm x_i-\bm x_j\rangle  \leq 0\quad  i,j \in [n]\times[n].\notag
\end{align}
Then, we estimate $f(\bm x)$ via
\begin{equation}\label{eqn:plc_estimate}
\hat f(\bm x) \triangleq \max_i  \langle \bm a_i, \bm x - \bm x_i \rangle + \hat y_i.
\end{equation}
This is a model similar to \citet{balazs2016convex} with a minor modification of using a different penalty term in the loss function. 
Our penalty term depends on the sum of partial derivatives $\sum_{l=1}^d \sup_{\bm x} | \partial_{x_l} f(\bm x)|$ rather than $\sup_{\bm x} \|\nabla f(\bm x)\|^2_2$. This new penalty term acts similar to a $L1$ regularizer and encourages feature sparsity. It is easy to show the estimator is bounded i.e., $\sup_{\bm x \in \Omega}|\hat f(\bm x)|  \leq M + 4 \|\hat f\|R$. Also $\|f+g\|\leq \|f\| +\|g \|$, $\|fc\|  =  c\|f\|$ for $c\geq0$.
Hence, similar to \citet{siahkamari2020piecewise}, our penalty term is a valid seminorm which allows us to use their theorem here.
\begin{proposition}\label{pro:convexregression}
With the appropriate choice of $\lambda$ which requires knowledge of $M$ the bound on $f$ and $n\ge d$, it holds that with probability at least $1-\delta$ over the data, the estimator $\hat{f}$ of $(\ref{eqn:plc_estimate})$ has excess risk upper bounded by 
\begin{align*} &\mathbb{E}[ |f(\bm x) - \hat{f}(\bm x)|^2]  \le O\bigg(  \bigg ( \frac{n}{d} \bigg )^\frac{-2}{d+4} \log \bigg (\frac{n}{d} \bigg ) +  \sqrt{\frac{\log(1/\delta)}{n}} \bigg) .\end{align*}
\end{proposition}

\subsection{Optimization}

Our method utilizes the well-known ADMM \citep{gabay1976dual} algorithm. ADMM is a standard tool for solving convex problems that consists of variable blocks with linear constraints \citep{eckstein2012augmented}. It has an iterative procedure to update the problem variables with provable convergence guarantees \citep{admm_convergence}. 

We solve program (\ref{program:convex}) using ADMM; the main insight is that we can solve a 2-block ADMM formulation where each block can be computed in closed form.  We first consider an equivalent form of the optimization problem as:
\begin{align} \label{program:convex_st}
&\qquad \min_{\substack{\hat{y}_i, \bm a_i, L_l\geq 0, \\ \bm p_i^+\geq 0 ,\bm p_i^- \geq 0,  \bm u_i \geq 0,  s_{i,j} \geq 0}}  \sum_{i=1}^n  (\hat y_i - y_i)^2 {+} \lambda \sum_{l=1}^d L_l\\
& \textrm{s.t.}\begin{cases}s_{i,j} + \hat y_i -  \hat y_j - \langle \bm a_i, \bm x_i-\bm x_j\rangle  = 0&  i,j \in [n]\times[n]  \\
u_{i,l} + p_{i,l}^+ + p_{i,l}^-  - L_l = 0 &\,i,l \in[n]\times[d]\\
a_{i,l}  -  p_{i,l}^+ + p_{i,l}^- =0 &\,i,l \in[n]\times[d],\notag
\end{cases}
\end{align}
with the augmented Lagrangian  \begin{align*}
&\ell(\hat{y}_i, \bm a_i, L_l, \bm p_i^+,\bm p_i^-, \bm u_i,  s_{i,j})= \frac{1}{n}\sum_{i = 1}^n  (\hat y_i - y_i)^2 + \lambda \sum_{l=1}^{d} L_l\notag\\
&+\sum_{i}\sum_{j}  \frac{\rho}{2}(s_{i,j} + \hat y_i -  \hat y_j - \langle \bm a_i, \bm x_i-\bm x_j\rangle+\alpha_{i,j})^2 \notag\\
&+\sum_{i}\sum_{l}  \frac{\rho}{2} (u_{i,l} + p_{i,l}^+ + p_{i,l}^- - L_l + \gamma_{i,l} )^2 \notag\\
 &+ \sum_{i}\sum_{l} \frac{\rho}{2}(a_{i,l}  -  p_{i,l}^+ + p_{i,l}^-+ \eta_{i,l})^2,
\end{align*}

where $\alpha_{i,j}$, $\gamma_{i,l}$ and $\eta_{i,j}$ are dual variables. We divide parameters into two blocks as ${\bm b}^1=\{\hat{y}_i, \bm a_i\}$ and ${\bm b}^2=\{L_l, \bm p_i^+, \bm p_i^-,\bm u_i,  s_{i,j}\}$, where ${i,j \in [n]\times[n], l\in [d]}$. We find closed form solutions for each block given the solution to the other block.

\subsubsection{First block ${\bm b}^1=\{\hat{y}_i, \bm a_i\}$}
We first note that we always normalize the dataset such that $\sum_i \bm x_i=\bm 0$ and $\sum_i y_i=0$. This will result in $\sum_i \hat y_i=0$ and simplify the solutions.

By setting $\nabla_{\bm a_i}\ell = \bm 0$ we can solve for $\bm a_i$ as:\begin{align}\label{sol:a}
    \bm a_i = \bm \Lambda_i( \bm \theta_i+ \hat  y_i \bm x_i + \frac{1}{n}\sum_k \hat y_k \bm x_k ),
  \end{align}
  where \begin{align*}
      \bm \Lambda_i &\triangleq (\bm x_i\bm x_i^T + \frac{1}{n}I + \frac{1}{n}\sum_j\bm x_j\bm x_j^T )^{-1},\notag\\
      \bm \theta_i &\triangleq  \frac{1}{n} \bigg(\bm p_{i}^+ - \bm p_{i}^- - \bm \eta_{i}+ \sum_j(\alpha_{i,j} + s_{i,j})(\bm x_i - \bm x_j)\bigg).
  \end{align*} 

Similarly by setting $\partial_{\hat y_i}\ell=0$ and substituting Eq. (\ref{sol:a}) for $\bm a_i$
we can solve for $\hat y_1, \dots, \hat y_n$, simultaneously as a system of linear equations,
\begin{align}\label{sol:y}
    \hat {\bm y} &= \bm \Omega^{-1}  \bigg(\frac{2\bm y}{n^2\rho} + \bm v -\bm \beta \bigg)
    \end{align} 
    where ${\bm y} = [ {y}_1, \dots,  {y}_n ]^T$,  $\hat {\bm y} = [\hat {y}_1, \dots, \hat {y}_n ]^T$, and
\begin{align*}
\beta_i &\triangleq \frac{1}{n} \sum_j\alpha_{i,j} - \alpha_{j,i} + s_{i,j} - s_{j,i},  \notag\\
     v_i &\triangleq    \bm x_i^T  \bm \Lambda_i \bm \theta_i  + \bm x_i^T \frac{1}{n} \sum_j \bm \Lambda_j \bm \theta_j-\frac{1}{n} \sum_j  \bm x_j^T\bm \Lambda_j \bm \theta_j\notag\\
         \Omega_{i,j} &\triangleq \bigg(\frac{2}{ n^2 \rho }+2- \bm x_i^T \bm \Lambda_i  \bm x_i\bigg )\mathbbm{1}(i=j)- \frac{1}{n}D_{i,j}, \notag\\
     D_{i,j} &\triangleq \bm x_i^T \bigg(\bm \Lambda_i\!+\!\bm \Lambda_j \!+\! \frac{1}{n}\!\sum_k\! \bm \Lambda_k\bigg) \bm x_j - \bm x_j^T\bm \Lambda_j \bm x_j \notag\\&- \frac{1}{n}\sum_k\bm x_k \bm \Lambda_k\bm x_j. 
     \end{align*}

\subsubsection{Second block ${\bm b}^2=\{L_l , \bm p_i^+, \bm p_i^-, \bm u_i,  \}$}\label{sec:2ndblock}
Set  $\partial_{s} \ell= 0$ for $s\in\{s_{i,j},\bm p_i^+, \bm p_i^-, \bm u_i, \}$. Hence \begin{align}\label{sol:s}
 s_{i,j} &= (-\alpha_{i,j}- \hat y_i +  \hat y_j + \langle \bm a_i, \bm x_i-\bm x_j\rangle )^+,\\
     u_{i,l} &= (L_l -\gamma_{i,l}  -|\eta_{i,l}+a_{i,l}| )^+,\notag\\
    p_{i,l}^+&=\frac{1}{2}(L_l  - \gamma_{i,l}- u_{i,l}+\eta_{i,l} + a_{i,l})^+,\notag\\
    p_{i,l}^-&=\frac{1}{2}(L_l  - \gamma_{i,l}- u_{i,l}-\eta_{i,l} - a_{i,l})^-.\notag
    \end{align} 
    
Lastly set $\partial_{L_l} \ell=0$ and 
plug the solutions for $ u_{i,l}$, $ p_{i,l}^+$ and $ p_{i,l}^-$. Denote $c_{i,l} \triangleq |\eta_{i,l} + a_{i,l}|$, after rearrangement of the terms we have:
\begin{align*}
 \frac{\lambda}{\rho} = \sum_i
\left\{
	\begin{array}{ll}
		0  & \mbox{if } L_l -\gamma_{i,l}  \geq c_{i,l} \\
		\frac{1}{2}(\gamma_{i,l}+ c_{i,l}-L_l) &  \mbox{if } | L_l -\gamma_{i,l}|  \leq c_{i,l} \\
		  \gamma_{i,l} - L_l & \mbox{if } L_l -\gamma_{i,l}  \leq - c_{i,l}.
	\end{array}
\right.
\end{align*}

Note that the right hand side is a monotonic and piecewise linear function of $L_l$. It is easy to find the solution to this problem with respect to $L_l$, using a sort and a simple algorithm that takes $O(n\log n)$ flops; see \textbf{L\_update} Algorithm \ref{algo:L_update}.
Observe that it is possible to add monotonicity constraints for $\hat f$ by projecting either $p^+_{i,l}$ or $p^-_{i,l}$ to zero. 
\subsubsection{Dual variables}
The update for dual variables follows from the standard ADMM algorithm updates:
\begin{align}\label{sol:dual}
& \alpha_{i,j} = \alpha_{i,j} + s_{i,j} \\
&+ \hat y_i - \hat y_j  - \langle \bm a_i, \bm x_i-\bm x_j\rangle & i,j \in [n]\times[n] \notag\\
& \gamma_{i,l} = \gamma_{i,l} + u_{i,l} + p_{i,l}^+ + p_{i,l}^-   - L_l  & i,l\in[n]\times[d] \notag \\
& \eta_{i,l} =  \eta_{i,l} + a_{i,l}  -  p_{i,l}^+ + p_{i,l}^-   & i,l\in[n]\times[d].\notag
\end{align}


\subsubsection{Algorithms via ADMM}\label{appx:admm_algor}

Algorithm \ref{code:convex} provides the full steps for a parallel ADMM optimizer for the convex regression problem. In each line in Algorithm~\ref{code:convex}, we use subscripts such as $i$, $j$ and $l$ on the left hand side. These updates may be run in parallel for $i\in[n]$, $i,j\in[n]\times[n]$, $i,l\in [n]\times[d]$ or $q\in[2]$. We initially set all block variables to zero and normalize the dataset such that $\sum_i y_i=0$ and $\sum_i \bm x_i=0$. We have implemented our algorithm using \textit{PyTorch} \citep{paszke2019pytorch}, which benefits from this parallel structure when using a GPU. Our code, along with a built-in tuner for our hyperparameter $\lambda$ and $T$, is available on our GitHub repository \footnote{ \href{https://github.com/Siahkamari/Piecewise-linear-regression}{https://github.com/Siahkamari/Piecewise-linear-regression} \label{link:github}}. 

\begin{algorithm}[t]
    \caption{\textbf{L-update}}
    \label{code:Lupdate}
    \begin{algorithmic}[1]\label{algo:L_update}
    \REQUIRE $\{\gamma_i, c_i\}_{i=1}^n$, and $\rho/\lambda$ 
    \STATE{} ${knot_{2n}, \dots, knot_{1}} \leftarrow  sort \{ \gamma_i + c_i, \gamma_i - c_i\}_{i=1}^n$ 
    \STATE{} $f \leftarrow \lambda/\rho$
    \STATE{} $f' \leftarrow 0 $
    \FOR{$j=2$ {\bfseries to} $2n$}
        \STATE{} $f' \leftarrow f'+ \frac{1}{2}$
        \STATE{} $f \leftarrow f +  f' \cdot (knot_{j} - knot_{j-1})$
        \IF{$f \leq 0$}
        \RETURN{} $ \big(knot_{j} -\frac{f}{f'}\big)^+  $
        \ENDIF
    \ENDFOR
    \RETURN{} $\big(knot_{2n} - \frac{f}{n} \big)^+ $
    \end{algorithmic}
\end{algorithm}

\begin{algorithm}[t!]
    \caption{Convex regression}
    \label{code:convex}
    \begin{algorithmic}[1]
    \REQUIRE $\{(\bm x_i, y_i)\}_{i=1}^n$, $\rho$, $\lambda$, and $T$
    \STATE{} $\hat y_i  = s_{i,j}  = \alpha_{i,j}  \leftarrow 0  $
    \STATE{} $\bm L =\bm a_i =  \bm p_i = \bm u_{i} = \bm \eta_i = \bm \gamma_{i}   \leftarrow \bm 0_{d\times 1} $
    \FOR{$t=1$ {\bfseries to} $T$} 
    \STATE  \textbf{Update} $\hat {\bm y}$ by Eq.~(\ref{sol:y})
    \STATE  \textbf{Update} $\bm a_i$ by Eq.~(\ref{sol:a})
    \STATE $L_l\leftarrow \textbf{L\_update}(\{\gamma_{i,l},|\eta_{i,l}+a_{i,l}|\}_{i\in [n]}, \lambda/\rho) $
    \STATE \textbf{Update} $ u_{i,l}, p_{i,l}^+, p_{i,l}^-, s_{i,j}$ by Eq.~(\ref{sol:s})
    \STATE \textbf{Update} $ \alpha_{i,j}, \gamma_{i,l}, \eta_{i,l}$ by Eq.~(\ref{sol:dual})
    \ENDFOR
    \RETURN{} $f(\cdot) \triangleq \max_{i=1}^n (\langle \bm a_i,\cdot - \bm x_i \rangle + \hat y_i ) $
\end{algorithmic}
\end{algorithm}

\subsection{Analysis}


Our method has two sources of errors which are the error due to the ADMM procedure (Eq. \ref{program:convex_st}) and the error due to estimating the ground truth convex function (Eq. \ref{eqn:plc_estimate}). We characterize both errors based on ADMM convergence \citep{admm_convergence}.

\begin{theorem}\label{thm.admm}[from \citet{admm_convergence}]
Consider the separable convex optimization problem,
 \begin{align*}
     &\min_{{\bf b}^1 \in \mathcal{S}_1, {\bf b}^2 \in \mathcal{S}_2} \left[\psi({\bf b}^1, {\bf b}^2) = \psi_1({\bf b}^1)+\psi_2({\bf b}^2)\right]\\
     & \quad s.t: \ma{\bf b}^1+\mb{\bf b}^2+\bm b = \bm 0,
 \end{align*}
where (${\bf b}^1$, ${\bf b}^2$) are the block variables, ($\mathcal{S}_1$, $\mathcal{S}_2$) are convex sets that includes all zero vectors, ($\ma,\mb$) are the coefficient matrices and ${\bf b}$ is a constant vector. Let ${\bf b}^1_t$ and ${\bf b}^2_t$ be solutions at iteration $t$ of a two block ADMM procedure with learning rate $\rho$ where $({\bf b}^1_0, {\bf b}^2_0)$ are all zero vectors. Denote average of iterates as $(\tilde{\bf b}^1_T, \tilde{\bf b}^2_T)=\left(\frac{1}{T}\sum_{t=1}^T{\bf b}^1_t, \frac{1}{T}\sum_{t=1}^T{\bf b}^2_t\right)$. For all ${\bm \kappa}$ we have,
\begin{align*}
&\psi(\tilde{\bf b}^1_T, \tilde{\bf b}^2_T)-\psi({\bf b}^1_*, {\bf b}^2_*)  - {\bm \kappa}^T (\ma\tilde{\bf b}^1_T+\mb\tilde{\bf b}^2_T+\bm b)\\
&\leq \frac{1}{T} \left(\frac{\rho}{2} \|\mb {\bf b}^2_* \|^2  + \frac{1}{2\rho} \|{\bm \kappa}\|^2\right),
\end{align*}
where $({\bf b}_1^*, {\bf b}_2^*)$ are the optimal solutions.
\end{theorem}



We arranged Theorem \ref{thm.admm} such that it explicitly depends on the problem dependent constants such as $\|\mb {\bf b}^2_* \|^2$, the number of iterations, $T$, as well as the learning rate $\rho$. 
 A 
proof 
is provided in Appendix \ref{sec.proof}. 

Next, we present convergence analysis in terms of regularized MSE of the output of the ADMM algorithm~(\ref{code:convex}). 
This has the further benefit of finding an appropriate learning rate $\rho$ and a range for regularization coefficient $\lambda$ that minimizes the computational complexity. 

\begin{theorem}\label{thm.main}
Let $\{\hat y_{i}^t, \bm a_{i}^t\}_{i=1}^n$ be the output of Algorithm \ref{code:convex} at the $t^{th}$ iteration, $\tilde {y_i} \triangleq \frac{1}{T}\sum_{t=1}^T\hat y_{i}^t$ and $\tilde {\bm a_i} \triangleq \frac{1}{T}\sum_{t=1}^T \bm a_{i}^t$. Denote
$\tilde f_T(\bm x) \triangleq \max_i  \langle \tilde{\bm a_i}, \bm x - \bm x_i \rangle + \tilde{y_i}$.
Assume $\max_{i,l} |x_{i,l}|\leq 1$ and $\mathbbm{Var}(\{y_i\}_{i=1}^n)\leq 1$.  If we choose $\rho=\frac{\sqrt d \lambda^2}{n}$, for $\lambda \geq \frac{3}{\sqrt{2nd}}$ and $T \geq {n\sqrt{d}}$ we have:
\begin{align*}
    &\frac{1}{n}\sum_{i=1}^n (\tilde f_T(\bm x_i) {-} y_i)^2 {+} \lambda \|\tilde f_T\|\\
    &\leq \min_{\hat f \in 
    \mathcal{F}} \bigg(\frac{1}{n} \sum_{i=1}^n\big(\hat f(\bm x_i) - y_i \big)^2 + \lambda \|\hat f\| \bigg) +  \frac{6n\sqrt{d}}{T+1} .
\end{align*}
\end{theorem}
\vspace{-0.2cm}
\begin{corollary}\label{thm.main.cor} Our method needs $T = \frac{6 n\sqrt{d}}{\epsilon} $ iterations to achieve $\epsilon$ error. Each iteration requires $\mathcal O (n^2d +nd^2)$ flops operations. Prepossessing costs $\mathcal O (n d^3)$. Therefore the total computational complexity is $\mathcal O \bigg (\frac{n^3 d^{1.5}+n^2 d^{2.5}+n d^3}{\epsilon}\bigg)$.
\end{corollary}
We note that assumptions of Theorem \ref{thm.main} are simply satisfied by normalizing the training dataset. The main difficulty for the derivation of Theorem \ref{thm.main} is: $(\tilde{\bm a_i}, \tilde{y_i})$ might be violating the constraints in (\ref{program:convex_st}). This could result in $\tilde f_T(\bm x_i) \neq \tilde y_i$ and  $\|\tilde f_T \| \neq \sum_{l=1}^d \tilde L_l$.  Therefore, the main steps of the proof of Theorem \ref{thm.main} is to characterize and bound the effect of such constraint violations on our objective function.
Then we set $\bm \kappa$ in Theorem~\ref{thm.admm} in a way to cover for the effects of constraint violations. For a full proof, we refer to Appendix \ref{sec.proof2}.



\section{Approximating a Bregman Divergence}

We next consider the application of learning a Bregman divergence from supervision. This problem is a type of metric learning problem, where we are given supervision (pairs of similar/dissimilar pairs, or triplets consisting of relative comparisons amongst data points) and we aim to learn a task-specific distance or divergence measure from the data.  Classical metric learning methods typically learn a linear transformation of the data, corresponding to a Mahalanobis-type distance function,
\begin{displaymath}
(\bm{x}-\bm{y})^T M (\bm{x}-\bm{y}),
\end{displaymath}
where $M$ is a positive semi-definite matrix (note that this generalizes the squared Euclidean distance, where $M$ would be the identity matrix).
More generally, Mahalanobis distances are examples of Bregman divergences, which include other divergences such as the KL-divergence as special cases. 

Bregman divergences are parameterized by an underlying strictly convex function, sometimes called the \textit{convex generating function} of the divergence.
Recently, \citet{siahkamari2019learning} formulate learning a Bregman divergence as learning the underlying convex function parameterizing the divergence.  The resulting optimization problem they study is similar to convex regression problem Eq.~(\ref{program:convex_st}).  Here we will discuss an improvement to their approach  with a slightly different loss function and our Lasso penalty term. We then derive an algorithm using 2-block ADMM; the existing solver for this problem uses standard linear programming.  

In particular
suppose we observe the classification dataset $S_n = \{(\bm x_i, y_i)\}_{i=1}^n$, for $\bm x_i \in \mathbb R^d$ and $y_i$ is an integer.  Let  
\begin{align*}
D_{f}(\bm x_i, \bm x_j) = f(\bm x_i) - f(\bm x_j) - \langle  \nabla^* f(\bm x_j), \bm x_i - \bm x_j\rangle,
\end{align*}
be a Bregman divergence with convex generating function ${f}(\bm x) \in \mathcal{F}$.  Let the pairwise similarity loss be
\begin{align*}
\ell(D_{f}(\bm x_i,\bm x_j),y_i,y_j) &= \\
\mathbbm{1}\big[\mathbbm{1}[&y_i=y_j]=(D_{f}(\bm x_i,\bm x_j)\geq 1)\big].
\end{align*}
We estimate $f$ the underlying convex function of the Bregman divergence $D_f$ by,
\begin{align*}
\hat {f} \triangleq \arg \min_{f\in \mathcal{F}} \sum_{i = 1}^n\sum_{j = 1}^n \ell(D_{f}(\bm x_i,\bm x_j),y_i,y_j)  + \lambda \| f \|.
\end{align*}
We note that this approach can easily be generalized to other possible metric learning losses.

\vspace{-.2cm}
\subsection{Optimization}\label{appendix:Breg}
\vspace{-.2cm}
Now we provide an algorithm for the Bregman divergence learning based on 2-block ADMM. We solve for each block in closed form. We first re-formulate the optimization problem to standard form and introducing some auxiliary variables  we have:
\begin{align}\label{program:pbdlpbdl}
&\min_{\substack{z_i, \bm a_i, \zeta_{i,j}\geq 0, \\ L_l \geq 0 ,  \bm p_i^+ \geq 0, \bm p_i^+ \geq 0, \\ \bm u_i \geq 0, s_{i,j}\geq 0, t_{i,j}\geq 0}} \sum_{i = 1}^n\sum_{j = 1}^n \zeta_{i,j} + \lambda \sum_{d}L_l\\
&\textrm{s.t.} \begin{cases}
\iota_{i,j}s_{i,j} -\iota_{i,j}  +t_{i,j}+1- \zeta_{i,j}=0 & i,j \in [n]\times[n] \\
s_{i,j}  + z_i - z_j - \langle \bm a_i, \bm x_i-\bm x_j\rangle  = 0 & i,j \in [n]\times[n] \\
p_{i,l}^+ + p_{i,l}^-  + u_{i,l}  - L_l = 0 & i,l\,\in[n]\times[d] \\
a_{i,l}  =  p_{i,l}^+ - p_{i,l}^- & i,l\, \in [n]\times[d],
\end{cases}\notag
\end{align}

where $\iota_{i,j} = 2(\mathbbm{1}[y_i = y_j]-\frac{1}{2})$.

Now we write the augmented Lagrangian:
\begin{align*}
&\ell( z_i, \bm a_i,\zeta_{i,j}, L_l, \bm p_i^+, \bm p_i^-, \bm u_i, s_{i,j}, t_{i,j})\\
&=   \sum_{i = 1}^n\sum_{j = 1}^n \zeta_{i,j} + \lambda \sum_{l=1}^{d} L_l\notag\\
&+ \sum_{i}\sum_{j} \frac{\rho}{2}(s_{i,j} + z_i -  z_j - \langle \bm a_i, \bm x_i-\bm x_j\rangle+ \alpha_{i,j})^2 \notag\\
&+ \sum_i\sum_l \frac{\rho}{2} (u_{i,l} + p_{i,l}^+ + p_{i,l}^- - L_l+\gamma_{i,l} )^2 \notag \\
& +\sum_{i}\sum_{l} \frac{\rho}{2}(a_{i,l}  -  p_{i,l}^+ + p_{i,l}^-+\eta_{i,l})^2 \notag \\
&+\sum_{i}\sum_{j}  \frac{\rho}{2}(\iota_{i,j}s_{i,j} -\iota_{i,j}  +t_{i,j}+1- \zeta_{i,j}+\tau_{i,j})^2,  \notag
\end{align*}

where $\alpha_{i,j}, \gamma_{i,j}, \eta_{i,j}$ and $\tau_{i,j}$ are dual variables. Next we divide the variables into two blocks and solve for each in closed form.

\subsubsection{First block ${\bm b}^1=\{ z_i, \bm a_i, \zeta_{i,j}\}$}
Setting $\partial_{\zeta_{i,j}} \ell =0$ gives:
\begin{align}\label{sol:zetam}
\zeta_{i,j} = (\frac{-1}{n\rho} + \tau_{i,j} + \iota_{i,j}s_{i,j} -\iota_{i,j}  +t_{i,j}+1)^+.
\end{align}
Collecting the terms containing $\bm a_i$ in Eq.(\ref{program:pbdlpbdl}) and comparing to those in Eq.~(\ref{program:convex_st}), the solution for $\bm a_i$ follows:

\begin{align}\label{sol:a2}
    \bm a_i \triangleq  \bm \Lambda_i( \bm \theta_i+  z_i \bm x_i + \frac{1}{n}\sum_k z_k \bm x_k ).
  \end{align}

Set $\partial_{z_i}\ell=0$, $\sum_i z_i=0$ and $\sum_i \bm x_i=\bm 0$. Using Eq.~(\ref{sol:a2})  we can solve for $z_1, \dots, z_n$ as
\begin{align}\label{sol:zm}
\bm z &= \bm \Omega_\text{breg}^{-1}(\bm \nu - \bm \beta),
\end{align}
where $\Omega_{\text{breg}_{i,j}} = (2-\bm x_i^T \Lambda_i \bm x_i)\mathbbm{1}[{i=j}]-\frac{1}{n}D_{i,j}$.

\subsubsection{Second block ${\bm b}^2=\{L_l, \bm p_i, \bm u_i, s_{i,j}, t_{i,j}\}$}
Comparing Eq.~(\ref{program:pbdlpbdl}) and Eq.~(\ref{program:convex_st}), we find that $ L_l$, $\bm p_i$ and $\bm u_i$, have the same solutions as in sec \ref{sec:2ndblock}. Hence we only proceed to solve for $s_{i,j}$ and $ t_{i,j}$.
Set $\partial_{t_{i,j}}\ell =\partial_{t_{i,j}}=0$. Some algebra gives:
\begin{align} \label{sol:tm}
 s_{i,j}&=  \frac{1}{2}(\pi^2_{i,j}+\iota_{i,j}\pi^1_{i,j}-\iota_{i,j}(\pi^1_{i,j}-\iota_{i,j}\pi^2_{i,j})^+)^+,\\
t_{i,j} &=(\pi^1_{i,j} -  \iota_{i,j}s_{i,j} )^+, \notag
\end{align}where \begin{align*}
    \pi^1_{i,j}  &\triangleq -\tau_{i,j}+\iota_{i,j}  -1+ \zeta_{i,j},\\
    \pi^2_{i,j} &\triangleq -\alpha_{i,j}-z_i + z_j + \langle \bm a_i, \bm x_i-\bm x_j\rangle.
\end{align*}

\subsubsection{Dual variables}
We only see a new constraint different from Eq.~(\ref{program:convex_st}) with dual multiplier $\tau_{i,j}$. The update is
\begin{align}\label{sol:tau}
    \tau_{i,j} &= \tau_{i,j} + \iota_{i,j}s_{i,j} -\iota_{i,j}  +t_{i,j}+1- \zeta_{i,j}.
\end{align}

Algorithm~\ref{code:pbdl} in the appendix provides full-steps for solving the Bregman divergence learning problem.


\vspace{-.3cm}
\section{Difference of Convex (DC) Regression}
\vspace{-.2cm}

As a further example, we extend our convex regression solver to the difference of convex regression as studied in \citet{siahkamari2020piecewise}. DC functions are set of functions $f$ that can be represented as $f = \phi^1 - \phi^2$ for a choice of two \emph{convex} functions. DC functions are a very rich class---for instance, they are known to contain all $\mathcal{C}^2$ functions. DC regression has been studied in \citet{cui2018composite, siahkamari2020piecewise, bagirov2020robust}. We provide a 2-block ADMM solver for the difference of convex estimator proposed in \citet{siahkamari2020piecewise}, with a minor change of choosing a different penalty 
\begin{align*}
\|f\|\triangleq \inf_{\phi^1,\phi^2}& \sum_{q=1}^2 \sum_{l=1}^d \sup_{\bm x} | \partial_{x_l}^* \phi^q(\bm x)|\\
    & \textrm{s.t. } \phi_1, \phi_2 \text{ are convex, } \phi_1 - \phi_2 = f.
\end{align*}

In particular, consider a regression dataset $\{(y_i,\bm x_i)\}_{i=1}^n$ with $\bm x_i \in \mathbb R^{d}$ and $y_i\in \mathbb R$. Denote the class of difference of convex functions
\begin{align*}
    \mathcal{DC} \triangleq \{f: \mathbb{R}^d \to \mathbb{R}\ | \ f = \phi^1 - \phi^2, (\phi^1,\phi^2) \text{ are convex}\},
\end{align*}
 In order to estimate $f$, we minimize the penalized least square over the $\mathcal{DC}$ class: 
\begin{align*}
\hat f \triangleq \arg \min_{f\in \mathcal{DC}} \frac{1}{n}\sum_{i=1}^n(y_i -f(\bm{x}_i))^2 + \lambda \|f\|.
\end{align*}
 This results in a convex program:
\begin{align}\label{program:dc}
&\quad \min_{\hat y_i^q , \bm a_i^q}\frac{1}{n} \sum_{i=1}^n  (\hat y_i^1 - \hat y_i^2  - y_i)^2 + \lambda \sum_{q=1}^2\sum_{l=1}^d \max_{i=1}^n | a_{i,l}^q| \notag\\
&\textrm{s.t. }\begin{cases}
\hat y_i^1 -  \hat y_j^1 - \langle \bm a_i^1, \bm x_i-\bm x_j\rangle  \leq 0 \quad  i,j \in [n]\times[n],\\
\hat y_i^2 -  \hat y_j^2 - \langle \bm a_i^2, \bm x_i-\bm x_j\rangle  \leq 0 \quad  i,j \in [n]\times[n],
\end{cases}
\end{align}
and the estimator for $f$ is given as
\begin{align*}
\hat f(\bm x) \triangleq& \max_i  \langle \bm a_i^1, \bm x - \bm x_i \rangle + \hat y_i^1 \\
 & -\max_i  \langle \bm a_i^2, \bm x - \bm x_i \rangle + \hat y_i^2.
\end{align*}
We note that the linear constraint sets for $\{\hat y_i^1, \bm a_i^1\}$ and $\{\hat y_i^2, \bm a_i^2\}$ have no variable in common in the linear program (\ref{program:dc}). Fixing $\hat{\bm{y}}^2$ allows us to solve for $\hat{\bm{y}}^1$ using Eq. (\ref{sol:y}) and vice-versa.
    From there we can decouple $\hat {\bm y}^1$ and $\hat {\bm y}^2$ solutions by solving a system of linear equations,
    \begin{align}\label{sol:y1y2}
    \hat {\bm y}^q &{=} \frac{(-1)^{q+1}}{2}\bigg (\bm \Omega {+} \frac{2I}{n^2\rho} 
    \bigg )^{-1} \bigg (\frac{4\bm y}{n^2\rho}\bm {+}v^1 -\bm \beta^1 {-} \bm v^2 {+}\bm \beta^2 \bigg ) \notag \\
    &{+}\frac{1}{2} \bigg (\bm \Omega - \frac{2\bm I}{n^2\rho} \bigg )^{-1}(\bm v^1 -\bm \beta^1 + \bm v^2 -\bm \beta^2).
\end{align}
Algorithm~\ref{code:dc} in Appendix provides the full steps to solve the difference of convex regression problem.
    
\vspace{-.3cm}
\section{Experiments}
\vspace{-.2cm}
\begin{figure*}[!htb] 
\includegraphics[width = 1\textwidth]{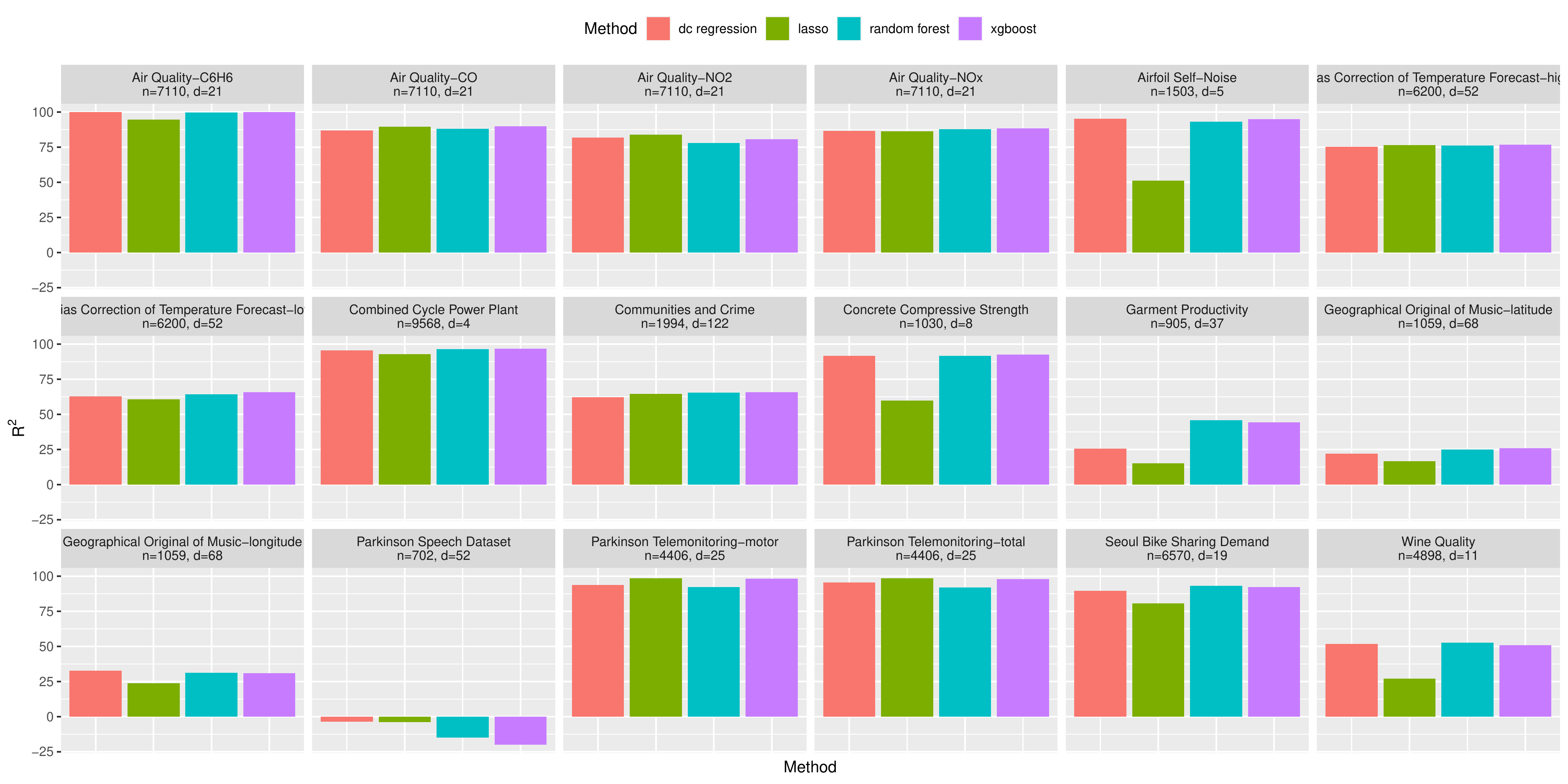}
\caption{Regression results on UCI datasets}\label{fig:2regression}
\end{figure*}
In this section we provide experiments on real datasets from the UCI machine learning repository as well as on synthetic datasets.  We compare running times between our proposed approach and the baseline interior point method for all three problems (convex regression, DC regression, and Bregman divergence learning).  We also compare both our DC regression algorithm as well as our Bregman divergence learning algorithm to state-of-the-art regression and classification methods, and show that our approach is close to state-of-the-art in terms of accuracy. We compare our (DC)-regression algorithm with state-of-the-art regression models. 

For the ADMM-based methods, we use a V100 Nvidia GPU processor with $11$ gigabyte of GPU memory, and 4 cores of CPU. For all the other methods we use a 16 core CPU.
\begin{table}[!t]
\centering
\small
\caption{Comparison of Convex Regression run time against baseline on Synthetic Data}
\begin{tabular}{ |c|c||c|c| } 
\hline
& & \multicolumn{2}{c|}{Seconds} \\
\hline
$n$ & $d$ & Baseline \cite{siahkamari2020piecewise} & This Paper \\
\hline
1000 & 2  & 32.3  & 3.63 \\
1000 & 4  & 30.8  & 3.75 \\
1000 & 8  & 54.5  & 4.03 \\
1000 & 16 & 112.3 & 4.12 \\
1000 & 32 & 189.3 & 4.37 \\
\hline
\end{tabular}
\label{tab:synth1}
\end{table}

\begin{table}[!t]
\centering
\small
\caption{Comparison of DC Regression run time against baseline on Synthetic Data}
\begin{tabular}{ |c|c||c|c| } 
\hline
& & \multicolumn{2}{c|}{Seconds} \\
\hline
$n$ & $d$ & Baseline \cite{siahkamari2020piecewise} & This Paper \\
\hline
1000 & 2 & 140.9 & 3.67 \\
1000 & 4 & 113.9 & 3.79 \\
1000 & 8 & 210.3 & 4.05 \\
1000 & 16 & 351.6 & 4.15 \\
1000 & 32 & 823.2 & 4.42 \\
\hline
\end{tabular}
\label{tab:synth2}
\end{table}

\vspace{-.2cm}
\subsection{Timing Results for Convex and DC Regression}
\vspace{-.2cm}
To demonstrate the effectiveness of our proposed approach, we first compare on synthetic data the existing interior point method to our 2-block ADMM method on standard convex regression and DC regression.  Here, we fixed the number of data points at 1000 and tested data of dimensionality $d \in \{2, 4, 8, 12, 32
\}$.

Results are shown in Tables~\ref{tab:synth1} and~\ref{tab:synth2}.  We see that our method ranges from 8.8x faster than the interior point method to up to 186x faster.  Note that the interior point method fails on large data sets, and cannot be run for many of the UCI data sets tested later.

\vspace{-.2cm}
\subsection{Results on Real Data}
\vspace{-.2cm}
Now we experiment with our DC regression algorithm and the Bregman divergence learning algorithm (PBDL) on real data sets, and report accuracy and timing results.  For PBDL, we compare against its predecessor (PBDL\_0) as well as state of the art classification models. All results are reported either on the pre-specifed train/test split or a 5-fold cross validation set based on instruction for each dataset. 

For the purpose of comparisons, we compare to the previous (DC)-regression and PBDL algorithms. Furthermore, we pick two popular tree-based algorithms XGboost and Random Forest. We also consider Lasso as a baseline. According to  \citet{fernandez2014we, fernandez2019extensive}, Random Forest achieves the state of the art on most regression and classification tasks. More recently, XGboost has been shown to achive the state of the art on many real datasets and \textit{Kaggle.com} competitions \cite{chen2016xgboost}.
However, we stress that achieving state-of-the-art results is not the purpose of our experiments. 

We provide the details of hyper-parameters for each method.

\textbf{DC-regression}: We choose $\lambda$ from a grid $10^{-3:3}$ by 5 fold cross validation. Then we do at most 2 more rounds of grid search around the optimal $\lambda$ at the first round. We fix $\rho=0.01$ and choose $T$ by early stopping, i.e., whenever validation error improvement is less than $10^{-3}$ after $n$ iterations of ADMM.
    
\textbf{PBDL:} We predict the classes using a  $5$ nearest neighbour scheme $i_{\text{nearest}}(\bm x)=\arg\min_i D_{\phi} (\bm x, \bm x_i)$. where $D_{\phi}$ is learned Bregman divergence. Tuning $\lambda$ is similar to DC-regression.
    
\textbf{PBDL\_0:} We use the existing code from the authors for learning a Bregman divergence based on Groubi solvers.
We use their built-in parameter tuner for $\lambda$ which is based on 5 fold cross validation.
    
\textbf{XGboost Regressor/Classifier:} $max\_depth$ and $learning\_rate$ is found by 5 fold cross validation over $\{1,\dots,10\}\times\{0.05,0.1,0.5\}$. Other parameters set to default. We use the xgboost 1.6.0 package in \citet{chen2016xgboost}.
    
\textbf{Random Forest Regressor/Classifier:}~$n\_features$ is found by 5 fold cross validation over $[d^{0.25},d^{0.75}] $. We use the sklearn package in \citet{scikit-learn}.

\textbf{Lasso:} We use the {sklearn} package function LassoCV.

\vspace{-.2cm}
\subsubsection{Datasets}
\vspace{-.2cm}
We chose all regression datasets from UCI which have number of instances $10^3\leq n\leq 10^4$. These were 30 datasets at the time of submission of this paper. We discard $16$ of these datasets due to various reasons such as the data set not being available.  Some of these datasets have multiple target variables, therefore we report the out of sample $R^2$ results with a suffix for each target variable. 

For classification, we chose all datasets originally used in \citet{siahkamari2019learning}. Further, we include the \textit{abalone} dataset which was too large for the previous PBDL algorithm. We use these datasets as multi-class classification problems.  We report accuracy as well as the training run-time.

\vspace{-.2cm}
\subsubsection{Results}
\vspace{-.2cm}
\textbf{Regression:} For our regression experiments we present the out of sample $R^2$ on $18$ datasets in figure \ref{fig:2regression} . We observe our method has similar performance to that of XGboost and Random Forest despite having $n\times d$ parameters. In solar-flare and Parkinson datasets, other methods overfit whereas our algorithm avoids overfitting and it is more robust.  Maximum number of instances which we were able to experiment on with reasonable time is about $10^4$ and is $10$x larger than \citet{siahkamari2020piecewise} has experimented on. The detailed results are in Appendix.

\begin{figure}[t] 
\includegraphics[width = 0.5\textwidth]{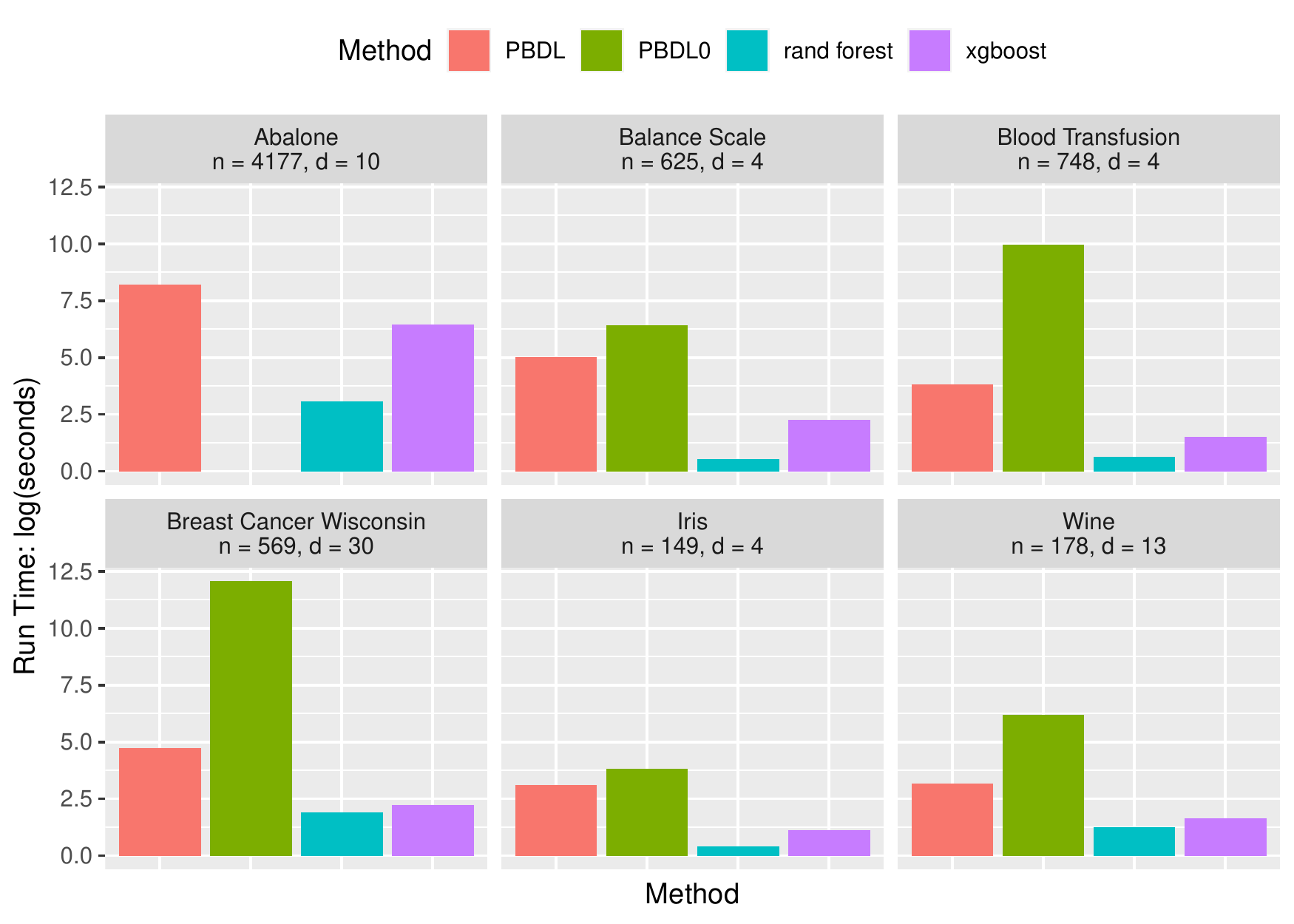}
\caption{Classification  log(runtime) for all fits on UCI datasets}\label{figure:class:acc}
\end{figure}

\textbf{Classification:} For our classification experiments we compare the logarithm of runtimes of all methods in Figure~\ref{figure:class:acc}. In terms of speed, we are on average $30$x faster than the original PBDL\_0 algorithm. Also PBDL\_0 fails to handle the Abalone dataset with $n=4177$, where the new PBDL takes less than a minute to finish one fit. We are slower than XGboost and Random Forest. However, we only rely on a python script code vs an optimized compiled code. We note that the divergence function learned in PBDL can be further used for other tasks such as ranking and clustering. We present the accuracy in Figure~\ref{figure:class:acc_app} in Appendix. We  observe that our Bregman divergence Learning Algorithm (PBDL) as well as the original (PBDL\_0) have similar performance to that of XGboost and Random Forest.

\section{Conclusion}
In this paper we,
studied the nonparametric convex regression problem with a $L1$ regularization penalty. we
 provided a solver for this problem and proved the iteration complexity be $6 n\sqrt{d}/\epsilon$. The total computational complexity is $O(n^3 d^{1.5}/\epsilon+n^2 d^{2.5}/\epsilon+n d^3/\epsilon)$, which improves that of $O(n^5 d^2/\epsilon)$ already known for this problem. We also extended our solver to the problem of difference of convex (DC) regression and the problem of learning an arbitrary Bregman divergence.
We provided comparisons to state of the art regression and classification models.



\section*{Acknowledgements}
This research was supported by the Secretary of Defense for Research and Engineering under Air Force Contract No. FA8702-15-D-0001, Army Research Office Grant W911NF2110246, the National Science Foundation grants CCF-2007350 and CCF-1955981, and the Hariri Data Science Faculty Fellowship Grants, and a gift from the ARM corporation. 

\bibliography{references}
\bibliographystyle{icml2022}


\onecolumn
\begin{appendix}
\thispagestyle{empty}
\begin{center}{ \textbf{\LARGE Appendix to `Faster Convex Lipschitz Regression via 2 blocks ADMM'}}\end{center}

\section{Algorithms}
\begin{algorithm}[ht]
   \caption{Learning a Bregman divergence}
   \label{code:pbdl}
\begin{algorithmic}[1]
\STATE {\bfseries Require:} $\{(\bm x_i, y_i)\,|\, \bm x_i \in \mathbb R^{d}, y_i \in \mathbb N\}_{i\in[n]}$ and  $\{\rho , \, \lambda, \, T\}$\\
    \STATE{} $z_i  = s_{i,j} = t_{i,j} = \alpha_{i,j} = \tau_{i,j} \leftarrow 0  $
    \STATE{} $\bm L =\bm a_i =  \bm p_i = \bm u_{i} = \bm \eta_i = \bm \gamma_{i}   \leftarrow \bm 0_{d\times 1} $
\FOR{$t=1$ {\bfseries to} $T$} 
\STATE \textbf{Update} $\zeta_{i,j}$ by Eq.~(\ref{sol:zetam})\\
\STATE \textbf{Update} $\bm z$ by Eq.~(\ref{sol:zm})\\
    \STATE  \textbf{Update} $\bm a_i$ by Eq.~(\ref{sol:a2})
\STATE $L_l\leftarrow \textbf{L\_update}(\{\gamma_{i,l},|\eta_{i,l}+a_{i,l}|\}_{i\in [n]}, \lambda/\rho) $\\
\STATE \textbf{Update} $ s_{i,j}, t_{i,j}$ by Eq.~(\ref{sol:tm})
\STATE \textbf{Update} $ u_{i,l}, p_{i,l}^+, p_{i,l}^-$ by Eq.~(\ref{sol:s})
\STATE \textbf{Update} $ \alpha_{i,j}, \gamma_{i,l}, \eta_{i,l}$ by Eq.~(\ref{sol:dual})
\STATE \textbf{Update} $ \tau_{i,j}$ by Eq.~(\ref{sol:tau})
\ENDFOR
\STATE {\bfseries return:} 
     $f(\cdot) \triangleq  \max_{i=1}^n (\langle \bm a_i,\cdot - \bm x_i  \rangle + z_i ) $
\end{algorithmic}
\end{algorithm}

\begin{algorithm}[ht]
    \caption{Difference of convex regression}
    \label{code:dc}
    \begin{algorithmic}[1]
    \REQUIRE $\{(\bm x_i, y_i)\}_{i=1}^n$, $\rho$, $\lambda$, and $T$
    \STATE{} $\hat y_i^q  = s_{i,j}^q  = \alpha_{i,j}^q  \leftarrow 0  $
    \STATE{} $\bm L^q =\bm a_i^q =  \bm p_i^q = \bm u_{i}^q = \bm \eta_i^q = \bm \gamma_{i}^q   \leftarrow \bm 0_{d\times 1} $
    \FOR{$t=1$ {\bfseries to} $T$} 
    \STATE  \textbf{Update} $\hat {\bm y}^q$  by Eq.~(\ref{sol:y1y2})
    \STATE  \textbf{Update} $\bm a_i^q$ by Eq.~(\ref{sol:a})
    \STATE $L_l^q\leftarrow \textbf{L\_update}(\{\gamma_{i,l}^q,|\eta_{i,l}^q+a_{i,l}^q|\}_{i\in [n]}, \lambda/\rho) $
    \STATE \textbf{Update} $ u_{i,l}^q, p_{i,l}^{q+}, p_{i,l}^{q-}, s_{i,j}^q$ by Eq.~(\ref{sol:s})
    \STATE \textbf{Update} $ \alpha_{i,j}^q, \gamma_{i,l}^q, \eta_{i,l}^q$ by Eq.~(\ref{sol:dual})

    \ENDFOR
    \RETURN{} $\phi^q(\cdot) \triangleq \max_{i=1}^n (\langle \bm a_i^q,\cdot - \bm x_i \rangle + \hat y_i^q ) $
\end{algorithmic}
\end{algorithm}

\section{Derivations}

\subsection{ADMM Convergence}\label{sec.proof}

Theorem \ref{thm.admm} gives convergence of the ADMM procedure. We follow analysis similar to \citet{admm_convergence} and give all steps for the sake of completeness. We first restate the convergence as follows: 
\setcounter{theorem}{0}
\begin{theorem}
Consider the separable convex optimization problem:
 \begin{align*}
     &\min_{{\bf b}^1 \in \mathcal{S}_1, {\bf b}^2 \in \mathcal{S}_2} \left[\psi({\bf b}^1, {\bf b}^2) = \psi_1({\bf b}^1)+\psi_2({\bf b}^2)\right]\\
     & \quad s.t: \ma{\bf b}^1+\mb{\bf b}^2+\bm b = \bm 0,
 \end{align*}
where (${\bf b}^1$, ${\bf b}^2$) are the block variables, ($\mathcal{S}_1$, $\mathcal{S}_2$) are convex sets that includes all zero vectors, ($\ma,\mb$) are the coefficient matrices and ${\bf b}$ is a constant vector. Let ${\bf b}^1_t$ and ${\bf b}^2_t$ be solutions at iteration $t$ of a two block ADMM procedure with learning rate $\rho$. i.e:
\begin{align*}
{\bf b}^1_{t+1}&=\arg\min_{{\bf b}^1\in \mathcal{S}_1} \psi_1({\bf b}^1)+\frac{\rho}{2}\big\|\ma {\bf b}^1 + \mb {\bf b}^2_t + {\bf b}-\frac{1}{\rho}{\bf d}_t\big\|^2\\
{\bf b}^2_{t+1}&=\arg\min_{{\bf b}^2\in \mathcal{S}_2} \psi_2({\bf b}^2)+\frac{\rho}{2}\big\|\ma {\bf b}^1_{t+1} + \mb {\bf b}^2 + {\bf b}-\frac{1}{\rho}{\bf d}_t\big\|^2\\
{\bf d}_{t+1}&={\bf d}_t-\rho\left(\ma {\bf b}^1_{t+1} + \mb {\bf b}^2_{t+1} + {\bf b}\right),
\end{align*}
where ${\bf b}^1_{0} = {\bf b}^2_{0} = {\bf d}_{0} = 0$. 

Denote the average of iterates as $(\tilde{\bf b}^1_T, \tilde{\bf b}^2_T)=\left(\frac{1}{T}\sum_{t=1}^T{\bf b}^1_t, \frac{1}{T}\sum_{t=1}^T{\bf b}^2_t\right)$. For all ${\bm \kappa}$ we have
\begin{align*}
\psi(\tilde{\bf b}^1_T, \tilde{\bf b}^2_T)-\psi({\bf b}^1_*, {\bf b}^2_*)  - {\bm \kappa}^T (\ma\tilde{\bf b}^1_T+\mb\tilde{\bf b}^2_T+\bm b)\leq \frac{1}{T} \left(\frac{\rho}{2} \|\mb {\bf b}^2_* \|^2 + \frac{1}{2\rho} \|{\bm \kappa}\|^2\right),
\end{align*}
where $({\bf b}_1^*, {\bf b}_2^*)$ are the optimal solutions.
\end{theorem}

Before presenting the proof, we define a dual variable like sequence $\{\overline{\bf d}_t\}_t$ with the update relation as 
$$\overline{\bf d}_{t+1}={\bf d}_t -\rho (\ma{\bf b}^1_{t+1}+\mb{\bf b}^2_{t}+\bf b).$$
The only difference between $\overline{\bf d}_{t+1}$ and ${\bf d}_{t+1}$ is due to ${\bf b}^2_{t}$ and ${\bf b}^2_{t+1}$. We have the following relation:
\begin{align}
    \overline{\bf d}_{t+1}-{\bf d}_{t+1}=\rho \mb ({\bf b}^2_{t+1}-{\bf b}^2_{t}).\label{eq.tilde}
\end{align}



 


 \begin{lemma}\label{lem.VI}[from \citet{nagurney1998network}]
Let ${\bf b}^*=\min_{{\bf b}\in\mathcal{S}}\psi({\bf b})$ where $\psi$ is continuously differentiable and $\mathcal{S}$ is closed and convex. Then $\bf b^*$ satisfies,
$$\left\langle \nabla \psi\left({\bf b}^*\right), {\bf b} - {\bf b}^*\right\rangle\geq 0,\quad \quad \forall {\bf b} \in \mathcal{S}.$$
\end{lemma}

Applying Lemma \ref{lem.VI} to optimization problems of ADMM gives,
\begin{flalign}
    \left\langle \nabla \psi\left({\bf b}^1_{t+1}\right) + \rho \ma^T \left(\ma {\bf b}^1_{t+1} + \mb {\bf b}^2_t + {\bf b}-\frac{1}{\rho}{\bf d}_t\right), {\bf b}^1 - {\bf b}^1_{t+1}\right\rangle\geq 0, \quad \forall {\bf b}^1 \in \mathcal{S}_1\label{pre_fo.1}\\
    \left\langle \nabla \psi\left({\bf b}^2_{t+1}\right) + \rho \mb^T \left(\ma {\bf b}^1_{t+1} + \mb {\bf b}^2_{t+1} + {\bf b}-\frac{1}{\rho}{\bf d}_t\right), {\bf b}^2 - {\bf b}^2_{t+1}\right\rangle\geq 0,\quad \forall{\bf b}^2 \in \mathcal{S}_2. \label{pre_fo.2}
\end{flalign}

Combining Eq. \ref{pre_fo.1}, \ref{pre_fo.2}, and the $\{\overline{\bf d}_t\}_t$ update rule we get, $\forall {\bf b}^1 \in \mathcal{S}_1, \forall{\bf b}^2 \in \mathcal{S}_2$,
\begin{flalign}
    \left\langle \nabla \psi\left({\bf b}^1_{t+1}\right)-\ma^T \overline{\bf d}_{t+1}, {\bf b}^1 - {\bf b}^1_{t+1}\right\rangle\geq 0,\quad \left\langle \nabla \psi\left({\bf b}^2_{t+1}\right) -\mb^T \overline{\bf d}_{t+1} - \rho\mb^T \mb ({\bf b}^2_{t}-{\bf b}^2_{t+1}), {\bf b}^2 - {\bf b}^2_{t+1}\right\rangle\geq 0. \label{fo}
\end{flalign}

Theorem \ref{thm.admm} is a direct conclusion of the following Lemma.

\begin{lemma}\label{lem.stepadmm}
For all ${\bf b}^1,{\bf b}^2, {\bf \kappa}$ we have, 
$$\psi({\bf b}^1,{\bf b}^2)-\psi({\bf b}^1_{t+1}, {\bf b}^2_{t+1}) + {\bm \kappa}^T (\ma{\bf b}^1_{t+1}+\mb{\bf b}^2_{t+1}+\bm b) - \overline{\bf d}_{t+1}^T(\ma {\bf b}^1 + \mb {\bf b}^2 + {\bf b}) +  { Z}_t - { Z}_{t+1} \geq  0,$$

where ${ Z}_t=\frac{\rho}{2} \|{\bf b}^2 - {\bf b}^2_{t}\|_{{\bf \mb^T \mb}}^2 + \frac{1}{2\rho} \|{\bm \kappa} - {\bf d}_t\|^2$ and $\|{\bf x}\|_{\bf M}^2=\bf x^T \bf M \bf x$.
\end{lemma}

Let ${\bf b}^1,{\bf b}^2$ be $({\bf b}_1^*, {\bf b}_2^*)$. By definition we have $\ma {\bf b}^1_* + \mb {\bf b}^2_* + {\bf b}={\bf 0}$ which cancels an inner product. Rearranging and averaging over time gives
\begin{align*}
    &\psi({\bf b}^1_{t+1}, {\bf b}^2_{t+1}) - \psi({\bf b}^1_*, {\bf b}^2_*)  - {\bm \kappa}^T (\ma{\bf b}^1_{t+1}+\mb{\bf b}^2_{t+1}+\bm b)  \leq { Z}_t -  { Z}_{t+1}\\
    &\frac{1}{T}\left(\sum_{t=0}^{T-1}\psi({\bf b}^1_{t+1}, {\bf b}^2_{t+1})\right) - \psi({\bf b}^1_*, {\bf b}^2_*) - {\bm \kappa}^T (\ma\tilde{\bf b}^1_{T}+\mb\tilde{\bf b}^2_{T}+\bm b)  \leq \frac{1}{T}\left(\sum_{t=0}^{T-1}{ Z}_t -  { Z}_{t+1}\right).
\end{align*}
The RHS telescopes. Since ${ Z}_t\geq 0$ we have $RHS\leq Z_0$. We lower bound LHS with Jensen Inq. as $\psi(\tilde{\bf b}^1_{T}, \tilde{\bf b}^2_{T})\leq \frac{1}{T}\sum_{t=0}^{T-1}\psi({\bf b}^1_{t+1}, {\bf b}^2_{t+1})$. Combining LHS and RHS we get
$$\psi(\tilde{\bf b}^1_{T}, \tilde{\bf b}^2_{T}) - \psi({\bf b}^1_*, {\bf b}^2_*) - {\bm \kappa}^T (\ma\tilde{\bf b}^1_{T}+\mb\tilde{\bf b}^2_{T}+\bm b) \leq \frac{1}{T}Z_0,$$
which is the statement in Theorem \ref{thm.admm} assuming the initial variables are $0$s.

We continue to prove Lemma \ref{lem.stepadmm}. We start with convexity definition for both $\psi_1$ and $\psi$ as,
\begin{align*}
    \psi_1({\bf b}^1) - \psi_1({\bf b}^1_{t+1}) - ({\bf b}^1 - {\bf b}^1_{t+1})^T \nabla \psi_1({\bf b}^1_{t+1}) \geq 0; \quad \psi_2({\bf b}^2) - \psi_2({\bf b}^2_{t+1}) - ({\bf b}^2 - {\bf b}^2_{t+1})^T \nabla \psi_2({\bf b}^2_{t+1}) \geq 0.
\end{align*}

Summing the relations and plugging in Eq. \ref{fo} give,
\begin{align}
    &\psi({\bf b}^1, {\bf b}^2) - \psi({\bf b}^1_{t+1}, {\bf b}^2_{t+1})\underbrace{ - ({\bf b}^1 - {\bf b}^1_{t+1})^T \ma^T \overline{\bf d}_{t+1} - ({\bf b}^2 - {\bf b}^2_{t+1})^T \mb^T \overline{\bf d}_{t+1} - \rho({\bf b}^2 - {\bf b}^2_{t+1})^T\mb^T \mb ({\bf b}^2_{t}-{\bf b}^2_{t+1})}_{T_1} \geq 0.\label{eq.convexity}
\end{align}

The update rule of $\overline{\bf d}$ gives $\ma{\bf b}^1_{t+1} + \mb{\bf b}^2_{t} +{\bf b}+ \frac{1}{\rho}(\overline{\bf d}_{t+1}-{\bf d}_t)= {\bf 0}$. Then for all ${\bm \kappa}$, we have $({\bm \kappa} - \overline{\bf d}_{t+1})^T(\ma{\bf b}^1_{t+1} + \mb{\bf b}^2_{t} +{\bf b}+ \frac{1}{\rho}(\overline{\bf d}_{t+1}-{\bf d}_t))=0$. Adding such a term to $T_1$ does not change its value.
After adding the zero inner product, we rearrange $T_1$ as
\begin{align*}
T_1=&\underbrace{{\bm \kappa}^T (\ma{\bf b}^1_{t+1}+\mb{\bf b}^2_{t+1}+\bm b)- \overline{\bf d}_{t+1}^T(\ma {\bf b}^1 + \mb {\bf b}^2 + {\bf b})}_{Y_1} + \rho\underbrace{({\bf b}^2 - {\bf b}^2_{t+1})^T\mb^T \mb ({\bf b}^2_{t+1}-{\bf b}^2_{t})}_{Y_2}\\
&+\underbrace{({\bm \kappa - \overline{\bf d}_{t+1}})^T(\mb({\bf b}^2_{t}-{\bf b}^2_{t+1})+\frac{1}{\rho}(\overline{\bf d}_{t+1}-{\bf d}_t))}_{Y_3}.
\end{align*}
Plugging Eq. \ref{eq.tilde} in $Y_3$ gives $Y_3= \frac{1}{\rho}({\bm \kappa - \overline{\bf d}_{t+1}})^T({\bf d}_{t+1}-{\bf d}_t)$.

We use the following norm square relation and prove it at the end,

\begin{lemma}\label{lem.norm}
For a symmetric ${\bf M}={\bf M}^T$ we have, 
$$({\bf x} - {\bf y})^T {\bf M} ({\bf z} - {\bf t}) = \frac{1}{2}\left( \|{\bf x} - {\bf t}\|_{{\bf M}}^2-\|{\bf x} - {\bf z}\|_{{\bf M}}^2\right)+\frac{1}{2}\left( \|{\bf y}-{\bf z}\|_{{\bf M}}^2-\|{\bf y} - {\bf t}\|_{{\bf M}}^2\right).$$
\end{lemma}

Using Lemma \ref{lem.norm}, we get

$Y_2 = \frac{1}{2}\left( \|{\bf b}^2 - {\bf b}^2_{t}\|_{{\bf \mb^T \mb}}^2-\|{\bf b}^2 - {\bf b}^2_{t+1}\|_{{\bf \mb^T \mb}}^2\right)-\frac{1}{2}\|{\bf b}^2_{t+1} - {\bf b}^2_{t}\|_{{\bf \mb^T \mb}}^2$

$Y_3=\frac{1}{\rho}\frac{1}{2}\left( \|{\bm \kappa} - {\bf d}_t\|_{{\bf I}}^2-\|{\bm \kappa} - {\bf d}_{t+1}\|_{{\bf I}}^2\right)+\frac{1}{\rho}\frac{1}{2}\left( \|\overline{\bf d}_{t+1}-{\bf d}_{t+1}\|_{{\bf I}}^2-\|\overline{\bf d}_{t+1} - {\bf d}_t\|_{{\bf I}}^2\right)$,

since ${\bf \mb^T \mb}$ and ${\bf I}$ are symmetric matrices.

We combine $Y_2$ and $Y_3$ as
\begin{align*}
    \rho Y_2 + Y_3=&\frac{\rho}{2}\left( \|{\bf b}^2 - {\bf b}^2_{t})\|_{{\bf \mb^T \mb}}^2-\|{\bf b}^2 - {\bf b}^2_{t+1}\|_{{\bf \mb^T \mb}}^2\right) + \frac{1}{2\rho}\left( \|{\bm \kappa} - {\bf d}_t\|_{{\bf I}}^2-\|{\bm \kappa} - {\bf d}_{t+1}\|_{{\bf I}}^2\right)\\
    &\underbrace{-\frac{1}{2\rho}\|\overline{\bf d}_{t+1}- {\bf d}_t\|_{{\bf I}}^2}_{\leq0}+
    \underbrace{\frac{1}{2\rho}\left( \|\overline{\bf d}_{t+1}-{\bf d}_{t+1}\|_{{\bf I}}^2-\rho^2\|{\bf b}^2_{t+1} - {\bf b}^2_{t}\|_{{\bf \mb^T \mb}}\right)}_{= 0 \text{ due to Eq. \ref{eq.tilde}}}\\
    \leq& \frac{\rho}{2}\left( \|{\bf b}^2 - {\bf b}^2_{t})\|_{{\bf \mb^T \mb}}^2-\|{\bf b}^2 - {\bf b}^2_{t+1}\|_{{\bf \mb^T \mb}}^2\right) + \frac{1}{2\rho}\left( \|{\bm \kappa} - {\bf d}_t\|_{{\bf I}}^2-\|{\bm \kappa} - {\bf d}_{t+1}\|_{{\bf I}}^2\right).
\end{align*}

Plugging it back to $T_1$,
\begin{align}
T_1=Y_1+\rho Y_2+Y_3\leq Y_1 + \frac{\rho}{2}\left( \|{\bf b}^2 - {\bf b}^2_{t})\|_{{\bf \mb^T \mb}}^2-\|{\bf b}^2 - {\bf b}^2_{t+1}\|_{{\bf \mb^T \mb}}^2\right) + \frac{1}{2\rho}\left( \|{\bm \kappa} - {\bf d}_t\|_{{\bf I}}^2-\|{\bm \kappa} - {\bf d}_{t+1}\|_{{\bf I}}^2\right)\leq Y_1 + { Z}_t - { Z}_{t+1}.\label{eq.last}
\end{align}

Upper bounding $T_1$ term in Eq. \ref{eq.convexity} with Eq. \ref{eq.last} gives
$$\psi({\bf b}^1,{\bf b}^2)-\psi({\bf b}^1_{t+1}, {\bf b}^2_{t+1}) + {\bm \kappa}^T (\ma{\bf b}^1_{t+1}+\mb{\bf b}^2_{t+1}+\bm b) - \overline{\bf d}_{t+1}^T(\ma {\bf b}^1 + \mb {\bf b}^2 + {\bf b}) +   { Z}_{t} -  { Z}_{t+1} \geq { 0},$$
which is the statement in Lemma \ref{lem.stepadmm}.$\square$

{\bf Proof of Lemma \ref{lem.VI} [from \citet{nagurney1998network}].}

Let $\phi({\bf t})=\psi({\bf b}^*+t({\bf b} - {\bf b}^*))$ for $t\in[0, 1]$. We know that $\phi$ achieves its minimum at $t=0$. Since $\mathcal{S}$ is convex and closed, we have, $0 \leq \left[\nabla \phi(t)\right]_{t=0}=\left\langle \nabla \psi\left({\bf b}^*\right), {\bf b} - {\bf b}^*\right\rangle. \square$

{\bf Proof of Lemma \ref{lem.norm}.}

Expand RHS as
\begin{align*}
RHS&=
\frac{1}{2}\left({\bf t}^T {\bf M} {\bf t}-2{\bf x}^T {\bf M} {\bf t}-{\bf z}^T {\bf M} {\bf z}+2{\bf x}^T {\bf M} {\bf z}\right)+
\frac{1}{2}\left({\bf z}^T {\bf M} {\bf z}-2{\bf y}^T {\bf M} {\bf z}-{\bf t}^T {\bf M} {\bf t}+2{\bf t}^T {\bf M} {\bf y}\right)\\
&=-{\bf x}^T {\bf M} {\bf t}+{\bf x}^T {\bf M} {\bf z}-{\bf y}^T {\bf M} {\bf z}+{\bf t}^T {\bf M} {\bf y}\\
&=({\bf x} - {\bf y})^T {\bf M} ({\bf z} - {\bf t}).
\end{align*}
We reach the $LHS$. $\square$

\subsection{Proof for Theorem 2 (computational complexity)}\label{sec.proof2}
\setcounter{theorem}{1}
\begin{theorem}
Let $\{\hat y_{i}^t, \bm a_{i}^t\}_{i=1}^n$ be the output of Algorithm(\ref{code:convex}) at the $t^{th}$ iteration, $\tilde {y_i} \triangleq \frac{1}{T}\sum_{t=1}^T\hat y_{i}^t$ and $\tilde {\bm a_i} \triangleq \frac{1}{T}\sum_{t=1}^T \bm a_{i}^t$. Denote
$\tilde f_T(\bm x) \triangleq \max_i  \langle \tilde{\bm a_i}, \bm x - \bm x_i \rangle + \tilde{y_i}$.
Assume $\max_{i,l} |x_{i,l}|\leq 1$ and $\mathbbm{Var}(\{y_i\}_{i=1}^n)\leq 1$.  If we choose $\rho=\frac{\sqrt d \lambda^2}{n}$, for $\lambda \geq \frac{3}{\sqrt{2nd}}$ and $T \geq {n\sqrt{d}}$ we have:
\begin{align*}
    \frac{1}{n}\sum_i (\tilde f_T(\bm x_i) {-} y_i)^2 {+} \lambda \|\tilde f_T\|\leq \min_{\hat f } \bigg(\frac{1}{n} \sum_i\big(\hat f(\bm x_i) - y_i \big)^2 + \lambda \|\hat f\| \bigg) + \frac{6n\sqrt{d}}{T+1} .
\end{align*}
\end{theorem}

 For the proof, we make extensive use of Theorem (\ref{thm.admm}) which provides convergence rate of a general 2-block ADMM for a separable convex program with linear constraints
 \begin{align}\label{program:general}
     &\min_{{\bf b}^1 \in \mathcal{S}_1, {\bf b}^2 \in \mathcal{S}_2} \left[\psi({\bf b}^1, {\bf b}^2) = \psi_1({\bf b}^1)+\psi_2({\bf b}^2)\right]\\
     & \quad s.t: \ma{\bf b}^1+\mb{\bf b}^2+\bm b = \bm 0. \notag
 \end{align}
It guarantees for all ${\bm \kappa}$ the average solutions of a 2-block ADMM $(\tilde{\bf b}^1_T, \tilde{\bf b}^2_T)$ satisfies
\begin{align}\label{thorem1:appendix}
\psi(\tilde{\bf b}^1_T, \tilde{\bf b}^2_T)-\psi({\bf b}^1_*, {\bf b}^2_*)  - {\bm \kappa}^T (\ma\tilde{\bf b}^1_T+\mb\tilde{\bf b}^2_T+\bm b)\leq \frac{1}{T} \left(\frac{\rho}{2} \|\mb {\bf b}^2_* \|^2 + \frac{1}{2\rho} \|{\bm \kappa}\|^2\right),
\end{align}
where $({\bf b}_1^*, {\bf b}_2^*)$ are the optimal solutions of program (\ref{program:general}).

However in Eq. (\ref{program:convex_st}) we are solving a specific version of program (\ref{program:general}) of the form
\begin{align} \label{program:convex_st_2}
&\min_{\hat{y}_i, \bm a_i, \bm p_i, L_l\geq 0, \bm u_i \geq 0,  s_{i,j} \geq 0} \left[\psi(\hat{y}_i, \bm a_i, \bm p_i, L_l, \bm u_i ,  s_{i,j}){=}\frac{1}{n}\sum_{i=1}^n  (\hat y_i - y_i)^2 {+} \lambda \sum_{l=1}^d L_l\right]\\
& \textrm{s.t.}\begin{cases}-s_{i,j} + \hat y_i -  \hat y_j - \langle \bm a_j, \bm x_i-\bm x_j\rangle  = 0&  i,j \in [n]\times[n]  \\
u_{i,l} + |p_{i,l}|  - L_l = 0 &\,i,l \in[n]\times[d]\\
a_{i,l}  -  p_{i,l} =0 &\,i,l \in[n]\times[d].
\end{cases}\nonumber
\end{align}

To match the notations between (\ref{program:convex_st_2}) and (\ref{program:general}) denote the optimal/average ADMM solutions to program (\ref{program:convex_st_2}) as
\begin{align}\label{proof:vars}
    {\bf b}^1_*& = [\hat y_1, \dots, \hat y_n, a_{1,1}, a_{1,2}, \dots,a_{n,d} ]^T\notag\\
    {\bf b}^2_* &= [L_1, \dots, L_d, p_{1,1}, \dots, p_{n,d}, u_{1,1},\dots, u_{n,d}, s_{1,1},\dots, s_{n,n}]^T \notag\\
    \tilde{\bf b}^1_T & = [\tilde y_1, \dots, \tilde y_n, \tilde a_{1,1}, \tilde a_{1,2}, \dots, \tilde a_{n,d} ]^T\notag\\
    \tilde{\bf b}^2_T &= [\tilde L_1, \dots, \tilde L_d, \tilde p_{1,1}, \dots, \tilde p_{n,d}, \tilde u_{1,1},\dots, \tilde u_{n,d}, \tilde s_{1,1},\dots, \tilde s_{n,n}]^T.
\end{align}
Therefore the separable losses for the average iterates of ADMM for program (\ref{program:convex_st_2}) are
\begin{align} \label{program:convex_st2}
&\psi_1(\tilde{\bf b}^1_T)=\frac{1}{n}\sum_{i=1}^n  (\tilde y_i - y_i)^2,  \\
&\psi_2(\tilde{\bf b}^2_T)=\lambda \sum_{l=1}^d \tilde L_l. \nonumber
\end{align}
Note that convergence rate for  $\psi_1(\tilde{\bf b}^1_T) +\psi_2(\tilde{\bf b}^2_T)$ is available by setting $\bm \kappa=0$ in (\ref{thorem1:appendix}). However this is not sufficient for convergence of our objective $\frac{1}{n}\sum_i (\tilde f_T(\bm x_i) {-} y_i)^2 {+} \lambda \|\tilde f_T\|$. The reason is $(\tilde{\bf b}^1_T, \tilde{\bf b}^2_T)$ might be violating the constraints in (\ref{program:convex_st_2}). This could result in $\tilde f_T(\bm x_i) \neq \tilde y_i$ and  $\|\tilde f_T \| \neq \sum_{l=1}^d \tilde L_l$.  Therefore main steps of the proof of Theorem (\ref{thm.main}) is to characterize and bound the effect of such constraint violations on our objective function.  Let us specify how much each linear constraint in program (\ref{program:convex_st_2}) is violated,
\begin{align}\label{equ:violate_cn}
\varepsilon^1_{i,j} &\triangleq -\tilde s_{i,j} + \tilde y_i -  \tilde y_j - \langle \tilde{\bm a_j}, \bm x_i-\bm x_j\rangle,    \\
\varepsilon^2_{i,l}&\triangleq \tilde u_{i,l} + |\tilde p_{i,l}|  - \tilde L_l,   \notag \\
\varepsilon^3_{i,l}&\triangleq \tilde a_{i,l}  -  \tilde p_{i,l}.   \notag
\end{align}

We break the proof to smaller parts by first providing some intermediate lemmas.

\begin{lemma}\label{lemma.00}
$\psi({\bf b}^1_*, {\bf b}^2_*) = \frac{1}{n}\sum_i(\hat y_i - y_i)^2 + \lambda \sum_l  L_l \leq 1$. 
\end{lemma}
\begin{proof}
Note that $\bm b^1 = \bm b^2 = 0 $ is a feasible solution and $(\hat y_i, L_l)$ is the optimal solution. Also algorithm \ref{code:convex} normalizes the dataset such that $\sum_{i=1}^n y_i =0$, therefore
\begin{align*}
 \psi({\bf b}^1_*, {\bf b}^2_*) &\leq \psi(\bm 0, \bm 0)\\
    &=\frac{1}{n}\sum_{i=1}^n y_i^2 = \mathbbm{Var}(\{y_i\}_{i=1}^n)\leq 1.
\end{align*}
\end{proof}

\begin{lemma}\label{lemma.b*}
$ \|\mb {\bf b}^2_* \|_2^2 \leq    18 \frac{n^2}{\lambda^2}$.
\end{lemma}

\begin{proof}
Since $p_{i,l}$ is a feasible solution we have $|p_{i,l}|\leq L_l$. We also have:
\begin{align*}
    0 \leq s_{i,j} &= \hat y_i -  \hat y_j - \langle \bm a_j, \bm x_i-\bm x_j\rangle & \text{(by constraints definitions)}\\
    &\leq \langle \bm a_i - \bm a_j, \bm x_i-\bm x_j\rangle & \text{(convexity)}\\
    &\leq 2 \sum_l L_l |x_{i,l}+x_{j,l}| & \text{($\max_i |a_{i,l}|\leq L_l)$}\\
    &\leq 4 \sum_l L_l & \text{($\max_i |x_{i,l}|\leq 1)$} .
\end{align*}

Arranging the constraints in (\ref{program:convex_st_2}) into $(n^2 + nd + nd)$ rows and separating the ${\bf b}^1_*,{\bf b}^2_*$ coefficients as $\ma$ and $\mb$ in order we have:
\begin{align*}
    &\ma {\bf b}^1_* + \mb {\bf b}^2_* = 0\\
    &\mb {\bf b}^2_* =[- s_{1,1}, \dots, - s_{n,n}, 0_{1,1},\dots,0_{n,d}, -p_{1,1}, \dots, -p_{n,d} ]^T.
\end{align*}
Taking the norm
\begin{align*}
\|\mb {\bf b}^2_* \|_2^2  &=  \sum_{i,j}s_{i,j}^2 + \sum_{i,l} p_{i,l}^2\\
&\leq   n^2 (4\sum_{l=1}^d L_l)^2 + n \sum_{l=1}^d L_l^2 \\
&\leq 18 \frac{n^2}{\lambda^2}. & \text{(Lemma \ref{lemma.00})}
\end{align*}
\end{proof}

\begin{lemma}\label{lem.a}
$0 \leq \tilde f_T(\bm x_i) -\tilde y_i \leq \max_j(\varepsilon^1_{i,j})$.
\end{lemma}
\begin{proof}
 \begin{align*}
    \tilde y_i \leq \tilde f_T(\bm x_i) &= \max_j \langle  \tilde{\bm a_j} , \bm x_i - \bm x_j \rangle + \tilde y_j & \text{(Definition)}\\
    &= \max_j(\tilde y_i - \tilde s_{i,j} + \varepsilon^1_{i,j}) & \text{($\varepsilon^1_{i,j}$ definition)}\\
    &\leq \tilde y_i + \max_j(\varepsilon^1_{i,j}). & \text{($\tilde s_{i,j} \geq 0$)}
\end{align*}
\end{proof}

\begin{lemma}\label{lemma.b}
\begin{align*}
  \frac{1}{n}\sum_i  \big(\tilde f_T(\bm x_i) - y_i \big)^2    \leq \frac{1}{n}\sum_i  (\tilde y_i - y_i )^2+  \frac{1}{n}\sum_i  \max_j(\varepsilon^1_{i,j})^2 +  2\sqrt{\frac{1}{n}\sum_i\max_j(\varepsilon^1_{i,j})^2} \sqrt{1+ \frac{1}{T} \frac{9\rho n^2}{\lambda^2}}.
\end{align*}
\end{lemma}
\begin{proof}
\begin{align*}
    \frac{1}{n}\sum_i  \big(\tilde f_T(\bm x_i) - y_i \big)^2 &=\frac{1}{n}\sum_i  \big(\tilde f_T(\bm x_i) -\tilde y_i +\tilde y_i - y_i \big)^2 \\
    &= \frac{1}{n}\sum_i  (\tilde y_i - y_i )^2 + \frac{1}{n}\sum_i  \big(\tilde f_T(\bm x_i)- \tilde y_i \big)^2 +  \frac{2}{n} \big(\tilde f_T(\bm x_i)- \tilde y_i \big)\big(\tilde y_i - y_i \big)\\
    &\leq \frac{1}{n}\sum_i  (\tilde y_i - y_i )^2  + \frac{1}{n}\sum_i  \max_j(\varepsilon^1_{i,j})^2 +  \frac{2}{n} |\max_j(\varepsilon^1_{i,j})|\big|\tilde y_i - y_i \big|. & \text{(Lemma \ref{lem.a})}
\end{align*}

Let us focus on the last term on the RHS. 
\begin{align*}
   \frac{\big( \frac{1}{n} \sum_i \big|\max_j(\varepsilon^1_{i,j})||\tilde y_i - y_i | \big )^2}{\frac{1}{n}\sum_i(\max_j\varepsilon^1_{i,j})^2}  
    &\leq  \frac{1}{n}\sum_i(\tilde y_i - y_i)^2  & \text{(Cauchy–Schwartz)}\\
    &\leq  \frac{1}{n}\sum_i(\tilde y_i - y_i)^2 + \lambda \sum_l \tilde L_l & \text{($\tilde L_l \geq 0$)} \\
    &\leq \frac{1}{n}\sum_i(\hat y_i - y_i)^2 + \lambda \sum_l L_l  + \frac{1}{T} \frac{\rho}{2} \|\mb {\bf b}^2_* \|^2 & \text{(Use Theorem \ref{thm.admm} with $\bm \kappa=0$)} \\
    &\leq 1+ \frac{1}{T} \frac{\rho}{2} \|\mb {\bf b}^2_* \|^2  &\text{(Lemma \ref{lemma.00})} \\
        &\leq 1+ \frac{1}{T} \frac{9\rho n^2}{\lambda^2}. &\text{(Lemma \ref{lemma.b*})} 
\end{align*}
\end{proof}
On the other hand for regularization term we have:
\begin{lemma}\label{lemma.hf}
   $\|\tilde f_T\|  \leq \sum_{l}|\tilde L_l| + \sum_l  \max_i |\varepsilon^2_{i,l}|+\max_i |\varepsilon^3_{i,l}|$.
\end{lemma}
\begin{proof}
\begin{align*}
    \|\tilde f_T\| &= \sum_{l}\max_i|\tilde{\bm a}_{i,l}| & \text{(definition)}\\
                    &= \sum_{l}\max_i|\tilde{\bm p}_{i,l}+\varepsilon^3_{i,l}| & \text{($\varepsilon^3_{i,l}$ definition)}\\
                    &\leq \sum_{l}|\tilde L_l|+\sum_l \max_i|\varepsilon^2_{i,l}|+\max_i|\varepsilon^3_{i,l}|. & \text{($\varepsilon^2_{i,l}$ definition)}
\end{align*}
\end{proof}

Next we combine the previous lemmas to prove Theorem \ref{thm.main}. Before proceeding lets incorporate the definitions of constraint violations Eq. (\ref{equ:violate_cn}) in ${\bm \kappa}^T (\ma\tilde{\bf b}^1_T+\mb\tilde{\bf b}^2_T+\bm b)$ to get:
\begin{align}\label{equ:violate_cn2}
   \bm \kappa^T (\ma\tilde{\bf b}^1_T+\mb\tilde{\bf b}^2_T+\bm b) =  {\bm \kappa^1}^T \bm \varepsilon^1  + {\bm \kappa^2}^T \bm \varepsilon^2 + {\bm \kappa^3}^T \bm \varepsilon^3,
\end{align}
where
\begin{align*}
& \bm \kappa^1 \triangleq [\kappa^1_{1,1},\dots ,\kappa^2_{n,n}]^T, \qquad \bm \kappa^2 \triangleq [\kappa^2_{1,1},\dots ,\kappa^2_{n,d}]^T, \qquad \bm \kappa^3 \triangleq [\kappa^3_{1,1},\dots ,\kappa^3_{n,d}]^T\\
&\bm \varepsilon^1 \triangleq [\varepsilon^1_{1,1},\dots ,\varepsilon^2_{n,n}]^T, \qquad \bm \varepsilon^2 \triangleq [\varepsilon^2_{1,1},\dots ,\ \varepsilon^2_{n,d}]^T, \qquad \bm \varepsilon^3 \triangleq [\varepsilon^3_{1,1},\dots ,\ \ \varepsilon^3_{n,d}]^T.
\end{align*}

Substituting (\ref{equ:violate_cn2}) in (\ref{thorem1:appendix}) we have:
\begin{align}\label{equ:Delta}
\Delta &\triangleq \psi(\tilde{\bf b}^1_T, \tilde{\bf b}^2_T)-\psi({\bf b}^1_*, {\bf b}^2_*)  - {\bm \kappa^1}^T \bm \varepsilon^1  - {\bm \kappa^2}^T \bm \varepsilon^2 - {\bm \kappa^3}^T \bm \varepsilon^3\notag \\
&\leq \frac{1}{T} \left(\frac{\rho}{2} \|\mb {\bf b}^2_* \|^2 + \frac{1}{2\rho} \|{\bm \kappa^1}\|^2 + \frac{1}{2\rho} \|{\bm \kappa^2}\|^2 + \frac{1}{2\rho} \|{\bm \kappa^3}\|^2\right).
\end{align}
\begin{proof}
Add up both sides of Lemma \ref{lemma.hf} and Lemma \ref{lemma.b} and subtract $\psi({\bf b}^1_*, {\bf b}^2_*)$. The total approximation error due to ADMM optimization schema is $LHS$: 
\begin{align*}
LHS &\triangleq \frac{1}{n}\sum_i  \big(\tilde f_T(\bm x_i) - y_i \big)^2 + \lambda \|\tilde f_T\| -\min_{\hat f}\bigg(\frac{1}{n}\sum_i  \big(\hat f(\bm x_i) - y_i \big)^2 + \lambda \|\hat f\| \bigg ) \\
&\leq \psi(\tilde{\bf b}^1_T, \tilde{\bf b}^2_T)-\psi({\bf b}^1_*, {\bf b}^2_*) \\
&+\frac{1}{n}\sum_i  (\varepsilon^1_{i,l*} )^2 + 2\underbrace{\sqrt{\frac{1}{n}\sum_i(\max_j\varepsilon^1_{i,j} )^2}}_{\epsilon_1} \underbrace{ \sqrt{1+ \frac{1}{T} \frac{9\rho n^2}{\lambda^2}}}_{\epsilon_2} + \lambda \sum_l \big(\max_i |\varepsilon^2_{i,l}|+\max_i |\varepsilon^3_{i,l}|\big)\\
&\triangleq RHS.
\end{align*}

Assume $\epsilon_1 \leq \epsilon_2$ then,
\begin{align*}
    RHS &\leq \psi(\tilde{\bf b}^1_T, \tilde{\bf b}^2_T)-\psi({\bf b}^1_*, {\bf b}^2_*) +3\sqrt{\frac{1}{n}\sum_i(\max_j\varepsilon^1_{i,j} )^2}  \sqrt{1+ \frac{1}{T} \frac{9\rho n^2}{\lambda^2}}+ \lambda \sum_l \big(\max_i |\varepsilon^2_{i,l}|+\max_i |\varepsilon^3_{i,l}|\big).
\end{align*}

In Eq. (\ref{equ:Delta}) set 
\begin{align*}
\kappa^1_{i,j}&=
    \begin{cases}
    -\frac{3\sqrt{1+ \frac{1}{T} \frac{9\rho n^2}{\lambda^2}}  \varepsilon^1_{i,j}}{\sqrt{n\sum_i(\max_k \varepsilon^1_{i,k} )^2}} & \text{if } j = \arg \max_k \varepsilon^1_{i,k}\\
    0  & \text{otherwise }
    \end{cases}\\
\kappa^2_{i,l}&=
    \begin{cases}
    -\lambda \sign \varepsilon^2_{i,l} & \text{if } i = \arg \max_k |\varepsilon^2_{k,l}|\\
    0  & \text{otherwise }
    \end{cases}\\
\kappa^3_{i,l}&=
    \begin{cases}
    -\lambda \sign \varepsilon^3_{i,l} & \text{if } i = \arg \max_k |\varepsilon^3_{k,l}|\\
    0  & \text{otherwise.}
    \end{cases}
\end{align*}
We have
\begin{align*}
RHS = \Delta &\leq \frac{1}{T} \left(\frac{\rho}{2} \|\mb {\bf b}^2_* \|^2 + \frac{9}{2n\rho} (1+ \frac{1}{T} \frac{9\rho n^2}{\lambda^2}) + \frac{d}{\rho} \lambda ^2\right)\\
&\leq \frac{1}{T} \left(  \frac{9\rho n^2}{\lambda^2}  + \frac{9}{2n\rho} (1+ \frac{1}{T} \frac{9\rho n^2}{\lambda^2}) + \frac{d}{\rho} \lambda ^2\right).
\end{align*}

Set $\rho = \frac{\sqrt{d }\lambda^2}{n}$, and then we get: $LHS \leq RHS = \Delta \leq \frac{6n\sqrt{d}}{T}$ if $\lambda \geq \frac{3}{\sqrt{2nd}}$ and $n\rho T \geq \frac{9}{2}$.

Now assume $\epsilon_1 \geq \epsilon_2$. We get:
\begin{align*}
    RHS &\leq \psi(\tilde{\bf b}^1_T, \tilde{\bf b}^2_T)-\psi({\bf b}^1_*, {\bf b}^2_*) +3\frac{1}{n}\sum_i(\max_j \varepsilon^1_{i,j} )^2  + \lambda \sum_l \big(\max_i |\varepsilon^2_{i,l}|+\max_i |\varepsilon^3_{i,l}|\big).
\end{align*}

In Eq. (\ref{equ:Delta}) set 
\begin{align*}
\kappa^1_{i,j}&=
    \begin{cases}
  -\frac{4}{n}\varepsilon^1_{i,j}& \text{if } j = \arg \max_k \varepsilon^1_{i,k}\\
    0  & \text{otherwise }
    \end{cases}\\
\kappa^2_{i,l}&=
    \begin{cases}
    -\lambda \sign \varepsilon^2_{i,l} & \text{if } i = \arg \max_k |\varepsilon^2_{k,l}|\\
    0  & \text{otherwise }
    \end{cases}\\
\kappa^3_{i,l}&=
    \begin{cases}
    -\lambda \sign \varepsilon^3_{i,l} & \text{if } i = \arg \max_k |\varepsilon^3_{k,l}|\\
    0  & \text{otherwise.}
    \end{cases}
\end{align*}

We get
\begin{align*}
    \Delta &= \psi(\tilde{\bf b}^1_T, \tilde{\bf b}^2_T)-\psi({\bf b}^1_*, {\bf b}^2_*) +4\frac{1}{n}\sum_i(\max_j \varepsilon^1_{i,j} )^2  + \lambda \sum_l \big(\max_i |\varepsilon^2_{i,l}|+\max_i |\varepsilon^3_{i,l}|\big) \\
    &\leq \frac{1}{T} \left(  \frac{9\rho n^2}{\lambda^2} + \frac{16}{2\rho n^2}\sum_i(\max_j \varepsilon^1_{i,j} )^2 + \frac{d}{\rho} \lambda ^2\right).
\end{align*}
It's straightforward to see if $ n \rho T \geq 8$ we have
\begin{align*}
    LHS \leq RHS \leq \Delta -\frac{16}{2\rho n^2T}\sum_i(\max_j\varepsilon^1_{i,j} )^2 = O(\frac{1}{T}( \frac{\rho n^2}{\lambda^2}  + \frac{d}{\rho} \lambda ^2)).\end{align*}

Set $\rho = \frac{\sqrt{d }\lambda^2}{n}$ we get: $LHS \leq  \frac{3 n\sqrt{d}}{T}$. Only if $T\geq \frac{8}{\sqrt{d }\lambda^2 }$.
\end{proof}
\section{Experimental results}
This section provides the regression and classification experimental results in tabular format. Table 3 compares the regression performance of our method against baseline methods on benchmark datasets. Table 4 compares the classification accuracy of our method against baseline methods. We present the run time of our method on selected UCI datasets in Table 5. Here, we omit the baseline method because it did not run to completion on most of the datasets. Hyperparameters were fixed for all timing results.

\begin{table}[ht]
\centering
\small
\caption{Comparison of Regression performance on UCI datasets.}
\begin{tabular}{ |l||c|c||c|c|c|c| } 
  \hline
 &  &  & \multicolumn{4}{c||}{$R^2 \times 100$} \\
  \hline
dataset & $n$ & $d$ & \textbf{dc regression} & lasso & random forest & xgboost \\
  \hline
Parkinson Speech Dataset & 702 & 52 & -3.7 &  -4 & -14.8 & -20 \\
Garment Productivity & 905 & 37 & 25.5 & 15.2 & 45.9 & 44.2 \\
Concrete Compressive Strength & 1030 & 8 & 91.7 &  59.9 & 91.8 & 92.5 \\
Geographical Original of Music-1 & 1059 & 68 & 21.9 & 16.6 & 25.1 & 26 \\
Geographical Original of Music-2 & 1059 & 68 & 32.6 &  23.8 & 31.3 & 30.8 \\
Solar Flare-1 & 1066 & 23 & 3.3 & 6.1 & -28 & 6.2 \\
Solar Flare-2 & 1066 & 23 & -6.1 & -2.1 & -17 & 2.1 \\
Airfoil Self-Noise & 1503 & 5 & 95.2 & 51.1 & 93.3 & 95 \\
Communities and Crime & 1994 & 122 & 62.1 &  64.5 & 65.4 & 65.8 \\
SML2010 & 3000 & 24 & 90.8 &  96 & 93.4 & 92.1 \\
SML2010 & 3000 & 24 & 77.3 &  94 & 92.4 & 90 \\
Parkinson's Telemonitoring & 4406 & 25 & 93.8 &  98.4 & 92.3 & 98.2 \\
Parkinson's Telemonitoring & 4406 & 25 & 95.5 &  98.5 & 92 & 98 \\
Wine Quality & 4898 & 11 & 51.8 & 27.2 & 52.6 & 51 \\
Bias Correction of Temperature Forecast-1 & 6200 & 52 & 62.9  & 60.8 & 64.2 & 65.8 \\
Bias Correction of Temperature Forecast-2 & 6200 & 52 & 75.2  & 76.5 & 76.3 & 76.7 \\
Seoul Bike Sharing Demand & 6570 & 19 & 89.7 &  80.7 & 93.1 & 92.3 \\
Air Quality-1 & 7110 & 21 & 87 &  89.5 & 88.2 & 89.9 \\
Air Quality-2 & 7110 & 21 & 100 &  94.7 & 99.8 & 100 \\
Air Quality-3 & 7110 & 21 & 86.6 &  86.4 & 87.7 & 88.5 \\
Air Quality-4 & 7110 & 21 & 81.7 &  84 & 78.1 & 80.7 \\
Combined Cycle Power Plant & 9568 & 4 & 95.5 &  92.9 & 96.5 & 96.7 \\
 \hline
\end{tabular}
\end{table}

\begin{table}[ht]
\centering
\small
\caption{Comparison of Classification performance on UCI datasets.}
\begin{tabular}{ |l||c|c||c|c|c|c||c|c| } 
 \hline
 &  &  & \multicolumn{4}{c||}{Accuracy (\%)} & \multicolumn{2}{c|}{Run time of all fits (sec)} \\
 \hline
 dataset & $n$ & $d$ & \textbf{pbdl} & pbdl0 & rand forest & xgboost & \textbf{pbdl} & pbdl0 \\
 \hline

Iris & 149 & 4 & 96.0 & 98.7 & 96 & 96.6 & 22.1 & 46.2 \\
Wine    & 178 & 13 & 96.0 & 96.6 & 96.7 & 95.5 & 23.6 & 496.2 \\
Blood Transfusion & 748 & 4 & 72.6 & 74.8 & 72.9 & 78.4 & 46.3 & 21514.0 \\
Breast Cancer Wisconsin  & 569 & 30 & 94.4 & 96.5 & 96.1 & 96.0 & 114.0 & 224407 \\
Balance Scale & 625 & 4 & 84.6 & 93.0 & 83 & 92.2 & 150.3 & 617.0 \\
Abalone & 4177 & 10 & 22.6 & N/A & 24.9 & 26.2 & 3688.0 & N/A \\
 \hline
\end{tabular}
\end{table}

\begin{table}[ht]
\centering
\small
\caption{DC Regression run time on UCI datasets. (Baseline is ommitted because it does not run to completion on most datasets)}
\begin{tabular}{ |l||c|c||c| } 
\hline
dataset & $n$ & $d$ & This Paper (Seconds) \\
\hline
Parkinson Speech Dataset & 702 & 52 & 3.4 \\
Garment Productivity & 905 & 37 & 4.12 \\
Concrete Compressive Strength & 1030 & 8 & 3.28 \\
Geographical Original of Music & 1059 & 68 & 4.44 \\
Solar Flare & 1066 & 23 & 3.64 \\
Airfoil Self-Noise & 1503 & 5 & 4.92 \\
Communities and Crime & 1994 & 122 & 12.82 \\
SML2010 & 3000 & 24 & 21.78 \\
\hline
\end{tabular}
\end{table}

\begin{figure}[t] \centering
\includegraphics[width = 0.5\textwidth]{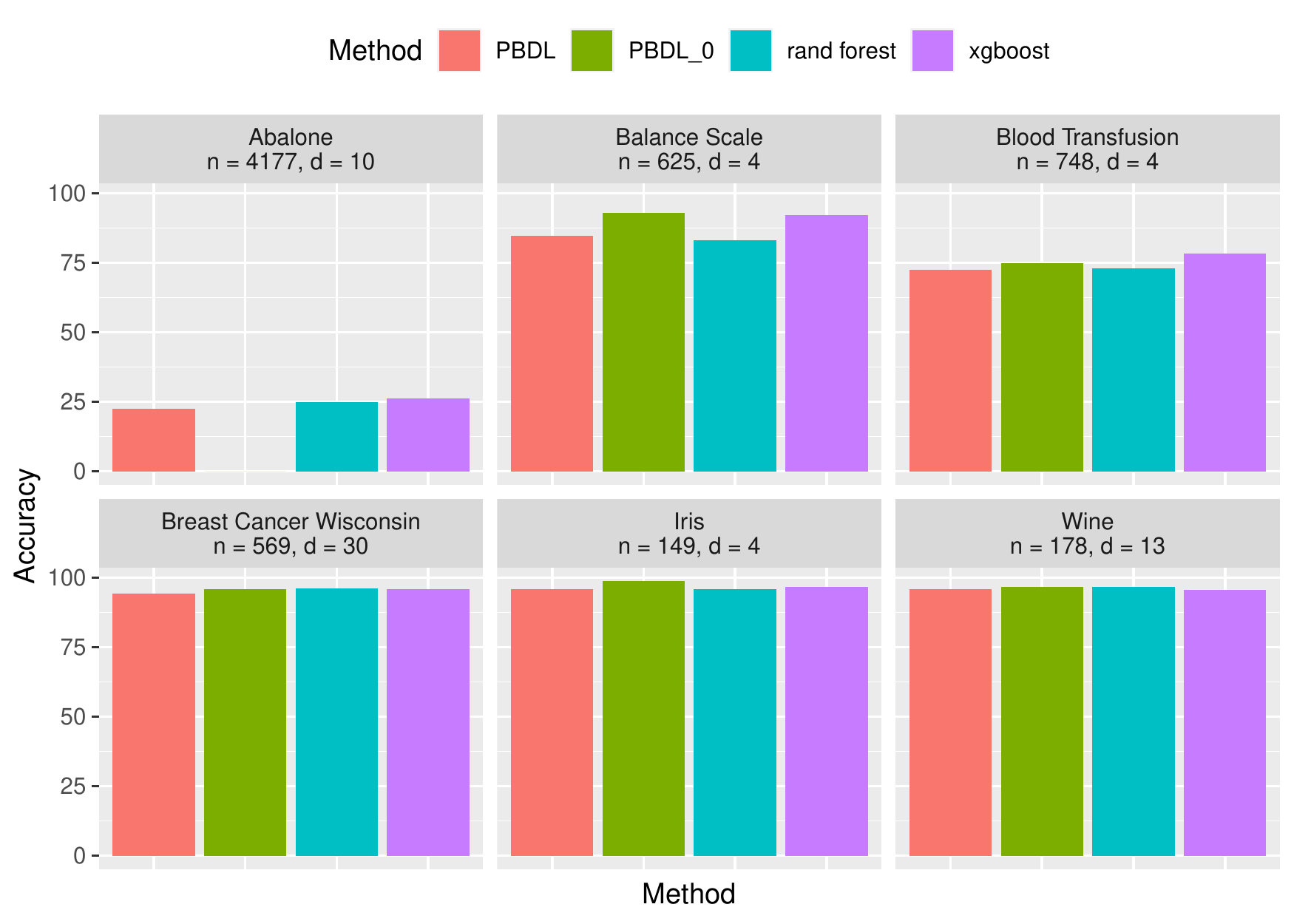}
\caption{Classification  accuracy for all fits on UCI datasets}\label{figure:class:acc_app}
\end{figure}

\end{appendix}

\end{document}